\documentclass[10pt,twocolumn,letterpaper]{article}

\usepackage[pagenumbers]{wacv} 

\usepackage{xcolor}
\usepackage{booktabs}
\usepackage{multirow}
\usepackage{amsmath}
\usepackage{amssymb}
\usepackage[table]{xcolor}
\usepackage{makecell}

\definecolor{metricgreen}{RGB}{0,120,0}

\definecolor{wacvblue}{rgb}{0.21,0.49,0.74}
\usepackage[pagebackref,breaklinks,colorlinks,allcolors=wacvblue]{hyperref}

\def\wacvPaperID{1450} 
\def\confName{WACV}
\def\confYear{2027}

\title{XGLASS: X-Ray, Gaze and Language Assisted Slice Sequencing for CT Volumes}

\author{
Michael Gl\"ockler and Timo Ropinski\\
Ulm University, Germany\\
{\tt\small \{michael.gloeckler, timo.ropinski\}@uni-ulm.de}
}

\begin{document}

\iftrue
\maketitle
\begin{abstract}
Vision-language models (VLMs) achieve remarkable performance on 2D medical images, such as X-rays or pathology images, while 3D VLMs that directly process CT/MRI volumes lag behind due to limited 3D training data and the predominance of 2D foundation architectures. To leverage powerful 2D VLMs for analyzing medical volumes, which contain hundreds of slices, it is therefore necessary to focus only on those few slices, which are diagnostically relevant. To enable such a prioritization, we propose XGLASS, a method that transfers medical expertise from 2D X-rays to 3D CT for task-aware slice prioritization. XGLASS uses disease-conditioned importance maps derived from language-aligned radiologist gaze on X-rays and spatially aligns them with coronal CT projections, bringing gaze priors and CT content into a shared 2D representation. The fused gaze-prior/CT projection is then used to predict a disease-specific ordering of axial slices, prioritizing diagnostically relevant slices. XGLASS significantly improves budgeted retrieval over 2D and 3D-native baselines, achieving a top-10 hit rate up to 75.9\% for the best-performing disease. In practice, this means that after inspecting only the top 10 slices (4.17\% of a 240-slice scan), XGLASS identifies a diagnostically relevant slice in more than 75\% of cases. For downstream slice-wise finding classification, an XGLASS-selected 30-slice candidate set reduces model calls by 87.5\% while improving F1 by 11.3 percentage points over all-slice inference, 4.2 points over the strongest 2D selector, and remains competitive with 3D-native selectors while requiring up to 130.8$\times$ less time and 10.9$\times$ less GPU memory per finding, supporting its use as a lightweight slice selector for volumetric CT analysis. The code and pretrained checkpoint are available on
\href{https://github.com/michaelgloeckler/XGLASS}{GitHub}.
\end{abstract}

\section{Introduction}
\label{sec:Introduction}
Vision-language models (VLMs) have recently enabled remarkable progress in computer vision, in tasks such as zero-/few-shot recognition, retrieval, and visual question answering (VQA), largely driven by the abundance of real-world image-text pairs~\cite{achiam2023gpt,alayrac2022flamingo,chen2022pali,team2025gemma,touvron2023llama,radford2021learning}. Analogously, the release of large paired medical datasets~\cite{irvin2019chexpert,johnson2019mimic,bustos2020padchest} has catalyzed medical adaptations, leading to CLIP-style medical contrastive models~\cite{wang2022medclip, lin2023pmc, zhang2022contrastive,zhang2025multimodal} and medical VLMs~\cite{myronenko2025reasoning,sellergren2025medgemma,li2023llava} capable of report generation, diagnosis, and medical VQA. Alongside these models, a growing ecosystem of 2D medical datasets has emerged, such as closed- and open-ended VQA benchmarks targeting medical questions~\cite{liu2021slake, he2020pathvqa,lau2018dataset,zhang2023pmc}. However, the majority of these resources contains 2D medical images, such as x-ray or pathology slides. When CT or MRI appear in those datasets, they are typically represented by single, diagnostically relevant slices that were manually extracted by experts, rather than complete 3D volumes~\cite{he2020pathvqa,lau2018dataset,liu2021slake}. This results in a mismatch between dataset availability and diagnostic requirements, since core diagnostic modalities, such as CT or MRI, are inherently 3D. Applying 2D VLMs directly to complete CT volumes would require processing every axial slice independently, which is computationally expensive and often inefficient because most slices are redundant or irrelevant to a queried finding. In practice, downstream applications such as slice-wise VLM screening, computer-aided detection (CAD), and 2D dataset curation benefit from compact disease-conditioned candidate sets rather than exhaustive all-slice inference~\cite{wang2024enhancing}. Motivated by this gap between predominantly 2D VLMs and the 3D nature of clinical imaging, we propose X-Ray, Gaze and Language Assisted Slice Sequencing (XGLASS), a lightweight method that transfers radiologist gaze-language priors learned on chest X-rays (CXRs) to rank axial CT slices by diagnostic relevance in a disease-conditioned manner. XGLASS effectively regresses a continuous, 240-dimensional relevance profile over axial slices, enabling relevance ranking, top-$k$ slice extraction, and importance-based traversal for retrieval or VLM conditioning. Empirically, XGLASS improves budgeted slice retrieval over 2D and 3D-native baselines, reaching 75.9\% Hit@10 for the best-performing disease, while downstream VLM screening shows that XGLASS-selected candidate sets reduce model calls by 87.5\% and improve F1 by up to 11.3 points over all-slice inference. \\To this end, we make the following contributions:
\begin{itemize}
  \item We introduce disease-conditioned importance maps from medical expert knowledge collected through x-ray eye-tracking data and synchronized speech recordings.
  \item We propose XGLASS, a lightweight 2D-to-3D cross-modal model that transfers these priors to CT projections and predicts disease-specific axial slice relevance scores.
  \item We evaluate XGLASS against 2D and 3D-native baselines, showing significantly better budgeted retrieval, substantially lower inference cost, and downstream F1 gains while reducing model calls by 87.5\%.
\end{itemize}

\section{Related Work}
\label{sec:RelatedWork}
\noindent\textbf{Eye-tracking in medical imaging.}
Radiologist eye-tracking provides implicit supervision of diagnostic search behavior and has been used to study visual expertise, diagnostic performance, and attention allocation~\cite{kundel2008using,van2017visual,drew2013scanners,kelahan2019radiologist}. Recent CXR datasets such as REFLACX and MIMIC-Eye link gaze with report transcripts~\cite{bigolin2022reflacx,hsieh2023mimic,Karargyris2020}, enabling gaze-guided localization, attention steering, and multimodal alignment~\cite{bigolin2023localization,ma2024eye,kim2024enhancing}. In contrast, eye-tracking data for 3D CT remain scarce: CT-Scangaze~\cite{pham2025ct} provides 3D scanpaths with dictation transcripts, but its limited number of only 57 distinct scanpath--report pairs makes it unsuitable for deriving stable disease-specific importance maps.

\begin{figure*}[t]
  \centering
  \includegraphics[width=\linewidth]{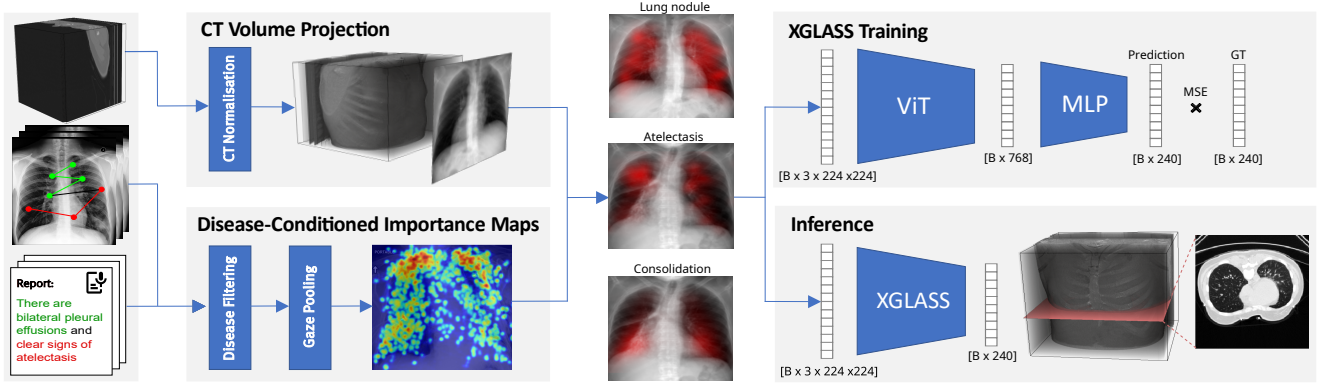}
\caption{\textbf{Overview of the XGLASS Pipeline.} 
Given a chest CT and chest X-ray eye-tracking data with matching report transcripts as input, we first normalize CT volumes and compute coronal CT intensity projections, while also matching gaze and disease mentions to derive disease-conditioned importance maps. These importance maps are spatially aligned to the CT projection and fused via alpha blending. The gaze-enriched intensity projection is input to the ViT-based regression head trained with MSE to predict a slice-importance vector. For inference, the model uses the precomputed importance map for the desired disease fused with the CT intensity projection and outputs a disease-specific relevance ordering of axial slices, ranking diagnostically important slices at the top.}
  \label{fig:pipeline}
\end{figure*}

\noindent\textbf{Slice selection for 3D medical imaging.}
CT and MRI interpretation requires scrolling through volumetric slices, making slice selection useful for reducing redundancy and enabling 2D models to process 3D data. Vote-MI~\cite{wang2024enhancing} selects representative slices from 2D features, but remains unsupervised, not disease-conditioned, and requires encoding all slices before selection, increasing computational cost. Although CT reading differs from CXR interpretation through slice drilling and z-axis-ordered search~\cite{drew2013scanners}, our CXR--CT gaze-prior comparison in Supplementary Sec.~\ref{sec:supp_cxr_ct_gaze_alignment} shows correlated superior--inferior disease-attention profiles across modalities ($r{=}0.65$--$0.96$), supporting XGLASS's transfer of 2D gaze priors for CT axial slice sequencing.

\section{Method}
\label{sec:Method}
Based on x-ray, gaze, and language input, XGLASS learns \(f(P, G) \rightarrow \hat{y}\), which processes a coronal CT intensity projection \(P\), enriched with a disease-specific gaze prior \(G\), in the form of an importance map. For clarity, the overall pipeline is illustrated in Figure~\ref{fig:pipeline}. Based on \(P\) and \(G\), \(f\) predicts a continuous importance vector \(\hat{y}\) of length \(Z_{\text{target}}\). Each entry of this vector represents the diagnostic relevance of a single axial slice with respect to the disease conditioned through the importance map. By learning \(f(P, G)\) with a ViT-based regression model, we enable effective disease-specific sequencing of the CT volume, from which \(P\) is obtained. In the following subsections, we first describe how to generate \(G\), then how to project a CT volume to obtain \(P\), and finally how we learn \(f\).

\subsection{Disease-Conditioned Importance Maps}
\label{sec:heatmap_gen}
A central component of XGLASS is the construction of disease-specific 2D importance maps \(G\) that capture how expert radiologists navigate through chest X-rays. To obtain these priors, we leverage the eye-tracking data paired with dictated report texts, available in the REFLACX dataset~\cite{bigolin2022reflacx}. For disease filtering for each report transcript, we use custom regular expressions similar to SARLE~\cite{draelos2021machine} to identify disease-specific mentions in the spoken text. The timestamps associated with these textual matches define a temporal window during which the radiologist was verbally describing the relevant findings. Prior work in psycholinguistics shows that speakers' gaze reliably anticipates verbal articulation by approximately one second, motivating the inclusion of a short pre-speech buffer on the order of 1–1.5\,s~\cite{griffin2000eyes}. Given the cognitive load and visual search demands of CT interpretation, we included a slightly longer pre-speech buffer of 2\,s to ensure we capture all relevant fixation points. Formally, the gaze pooling process is described as follows, for a spoken mention with start time $t_{\mathrm{start}}$ and end time $t_{\mathrm{end}}$, we define the effective gaze window as \([t_{\mathrm{start}} - 2\,\mathrm{s},\; t_{\mathrm{end}}]\). All gaze fixations that fall into this window are then extracted, ensuring that only gaze from the relevant diagnostic moment contributes to the disease-conditioned importance map. To obtain a consistent spatial representation, we normalize each fixation with respect to the REFLACX chest bounding box \(\mathcal{B}_{\text{chest}}=(x_0,y_0,w,h)\) and discard fixations falling outside \(\mathcal{B}_{\text{chest}}\) (potentially resulting from tracker drift or blinks). The box-normalized coordinates are

\[
\tilde{x} = \frac{x - x_0}{w}, \qquad
\tilde{y} = \frac{y - y_0}{h},
\]

\noindent so that \((\tilde{x},\tilde{y}) \in [0,1]^2\). We then aggregate the normalized fixations across studies for the same disease category and smooth them by placing a Gaussian at each fixation to obtain a continuous density map~\cite{wooding2002fixation}. Since REFLACX contains frontal CXRs acquired in both AP and PA views, we additionally analyze whether the resulting gaze priors are sensitive to view-specific acquisition geometry. In Sec.~\ref{sec:supp_ap_pa} of the supplementary material, AP- and PA-specific gaze maps show only minor differences in their superior--inferior profiles; therefore, we aggregate both views to maximize the number of gaze samples available per disease. The resulting disease-conditioned importance maps capture the aggregated spatial attention pattern for each finding and serve as weak expert annotation priors that are subsequently fused with coronal CT intensity projections to create the final input for XGLASS.

\subsection{CT Volume Projection}
\label{sec:volumeprojection}
Our method takes two complementary sources of supervision: pre-computed, disease-specific 2D gaze importance maps derived from radiologist eye-tracking on chest X-rays using the REFLACX dataset~\cite{bigolin2022reflacx}, and a volumetric CT dataset that shares matching disease findings and therefore ReXGroundingCT~\cite{baharoon2025rexgroundingct} is to our knowledge the largest CT dataset that fulfills our requirements. To project a given CT volume into the disease-conditioned importance map domain, we first need to perform an intensity projection followed by a projection window normalization.

\noindent\textbf{Intensity projection:} CT volume projections enable us to project a given 3D volume into a 2D image, which in case of a coronal projection, is aligned with our disease-conditioned importance maps. Formally, let $f(\mathbf{x})$ denote the volumetric attenuation at location $\mathbf{x}\in\mathbb{R}^3$. A projection along an imaging direction $\mathbf{d}$ samples parallel-beams $\mathbf{x}(u,v,t)=\mathbf{o}(u,v)+t\,\mathbf{d}$ and applies an aggregation operator $\mathcal{A}$ to the line signal $f(\mathbf{x}(u,v,t))$:

\begin{equation}
\label{eq:generic_projection}
\mathcal{P}_{\mathcal{A}}(u,v)\;=\;\mathcal{A}\big(\{\,f(\mathbf{o}(u,v)+t\,\mathbf{d})\,\}_{t\in[0,L]}\big).
\end{equation}

\noindent Because the coronal projection preserves the superior--inferior coordinate indexing axial slices, aggregating anterior--posterior depth retains the positional information required for axial slice selection. Accordingly, $\mathbf{d}$ follows the anterior--posterior axis, while different choices of $\mathcal{A}$ yield standard projection types. To simulate real-world x-ray projection $\mathcal{A_{BL}}$ would resemble the Beer-Lambert law~\cite{swinehart1962beer}, which assumes that x-rays are only attenuated but not reflected at $\mathbf{x}$:
\begin{equation}
\label{eq:abl}
\mathrm{A_{BL}}(u,v)\;=\;\exp(-\kappa\int_{0}^{L} f(\mathbf{o}(u,v)+t\,\mathbf{d})\,\mathrm{d}t),
\end{equation}
whereby \(\kappa\) is the constant extinction coefficient, which models absorption and out-scattering. An \emph{average-intensity projection} (AIP) in contrast uses the line integral normalized by its length:
\begin{equation}
\label{eq:aip}
\mathrm{A_{IP}}(u,v)\;=\;\frac{1}{L}\int_{0}^{L} f(\mathbf{o}(u,v)+t\,\mathbf{d})\,\mathrm{d}t.
\end{equation}
In the field of radiology, additionally, the \emph{maximum-intensity projection} (MIP), defined as $\mathrm{A_{MIP}}(u,v)=\max_{t\in[0,L]} f(\mathbf{o}(u,v)+t\,\mathbf{d})$, plays an important role. However, MIPs emphasize very high-density structures (e.g. bones) and de-emphasize soft-tissue pathology, which conflicts with the intensity ranges in which our target findings occur. Therefore, throughout this paper, we solely focus on the Beer-Lambert projection and the AIP. We have ablated both (see Section~\ref{sec:CTPablations}), whereby we compare three standard CT windowings using AIP, i.e., lung, soft-tissue, and a wide window.

\noindent\textbf{Projection window normalization.} To ensure the described intensity projections are aligned with the obtained disease-conditioned importance maps, anatomy normalization needs to be applied. During this normalization, our training set of 3,142 CT volumes was resampled to a uniform isotropic spacing of \(1.25 \times 1.25 \times 1.25\)\,mm. While this standardizes resolution, it does not reconcile variability in scan coverage along the \(Z\)-axis, which is critical since XGLASS 2D-to-3D matching expects a fixed amount of axial slices. To avoid the anatomical distortion introduced by global resizing (i.e., stretching or squashing), we adopt a fixed-size cropping strategy and define a target axial block size \(Z_{\text{target}}\) to be extracted from each volume. To further minimize exclusions, we sweep \(Z_{\text{target}}\) to find an optimal setting. A CT volume \(V_i\) is included only if two constraints hold: (1) its total axial slice count \(Z_{\text{total},i}\) (at 1.25\,mm) is large enough to contain the block, and (2) its pre-computed chest bounding-box height \(Z_{\text{bbox},i}\) (at 1.25\,mm) fits within the block.
\[
V_i \in \mathrm{Cohort}_{\mathrm{final}}
\;\Leftrightarrow\;
\big(Z_{\text{total},i} \ge Z_{\text{target}}\big)\;\wedge\;\big(Z_{\text{bbox},i} \le Z_{\text{target}}\big)
\]

We tested integer values \(Z_{\text{target}} \in [128, 512]\) and found an optimal value of \(Z_{\text{target}} = 240\) slices. This choice minimized exclusions by balancing chest-bbox inclusion (median \(Z_{\text{bbox},i}=205\)) against total volume length (median \(Z_{\text{total},i}=269\)). As a result, the final cohort comprises 2{,}078 unique CTs, with 1{,}064 excluded for not meeting the constraints. For all included CTs, we then extract a fixed axial block of 240 slices (1.25\,mm spacing) centered on the chest bbox, yielding a standardized ReXGroundingCT subset with a consistent 240-slice axial size across scans. Further dataset statistics and category-selection details are provided in Sec.~\ref{sec:supp_ct_dataset} of the supplementary material.

\noindent\textbf{Ground-truth slice-area vector extraction.}
\label{sec:data_prep_gt}
A key advantage of XGLASS is that it does not necessarily require full 3D segmentation masks: it can also be trained using only the start and end slices of a finding, as demonstrated by the SE-Binary and SE-Smooth ablations in Sec.~\ref{sec:weak_supervision}. When segmentation masks are available, as in ReXGroundingCT, we use the more detailed Slice-Area target by deriving per-slice relevance scores from the 3D ground-truth masks. Therefore each mask is converted into a 240-dimensional disease-area vector over the fixed axial block. Let \(A_i\) denote the disease area on slice \(i\), with \(A_i=0\) when the finding is absent. We compute
\[
M=\max_{j=1,\dots,240}A_j,
\]
and define the target vector \(\mathbf{y}=(y_1,\dots,y_{240})\) by max-normalization:
\[
y_i=
\begin{cases}
A_i/M, & \text{if } M>0,\\[4pt]
0, & \text{if } M=0,
\end{cases}
\qquad i=1,\dots,240.
\]
Thus, the slice with the largest disease extent receives \(y_i=1\), absent slices remain at \(0\), and intermediate slices receive values proportional to their disease area.

\newcommand{\metricci}[3]{\makecell[c]{#1\\[-0.25em]{\scriptsize [#2, #3]}}}
\newcommand{\bestmetricci}[3]{\makecell[c]{\textbf{#1}\\[-0.25em]{\scriptsize \textbf{[#2, #3]}}}}
\newcommand{\secondmetricci}[3]{\makecell[c]{\underline{#1}\\[-0.25em]{\scriptsize [#2, #3]}}}
\newcommand{\posdeltaci}[3]{\makecell[c]{\textcolor{metricgreen}{#1\%}\\[-0.00em]{\scriptsize \textcolor{metricgreen}{[#2, #3]}}}}
\newcommand{\negdeltaci}[3]{\makecell[c]{\textcolor{red}{#1\%}\\[-0.00em]{\scriptsize \textcolor{red}{[#2, #3]}}}}
\newcommand{\sigposdeltaci}[3]{\makecell[c]{\textcolor{metricgreen}{\textbf{#1\%}}\\[-0.00em]{\scriptsize \textcolor{metricgreen}{\textbf{[#2, #3]}}}}}

\newcommand{\posdelta}[1]{\textcolor{metricgreen}{#1\%}}
\newcommand{\negdelta}[1]{\textcolor{red}{#1\%}}
\newcommand{\sigposdelta}[1]{\textcolor{metricgreen}{\textbf{#1\%}}}

\begin{table*}[t!]
\centering
\small
\setlength{\tabcolsep}{3.8pt}
\renewcommand{\arraystretch}{1.15}
\caption{\textbf{Main Results.}
Metrics are reported as point estimate with 95\% confidence interval below. Intervals are computed using the nonparametric percentile bootstrap~\cite{efron1992bootstrap} over the test set with 10{,}000 resamples. The last rows report paired absolute differences in percentage points between XGLASS and the strongest 2D or 3D baseline for each metric; bold indicates statistical significance.}
\label{tab:main_results}
\begin{tabular*}{\textwidth}{@{\extracolsep{\fill}}lccccccc}
\toprule
\textbf{Method}
  & \textbf{Hit@1} & \textbf{Hit@3} & \textbf{Hit@5} & \textbf{Hit@10}
  & \textbf{AR@10} & \textbf{AUC} & \textbf{Spear. ($\rho$)} \\
\midrule
Random
& \metricci{0.101}{0.070}{0.136}
& \metricci{0.250}{0.203}{0.298}
& \secondmetricci{0.342}{0.291}{0.396}
& \secondmetricci{0.491}{0.437}{0.547}
& \metricci{0.043}{0.035}{0.051}
& \metricci{0.504}{0.497}{0.511}
& \metricci{0.003}{-0.004}{0.010} \\

Gaussian Centered
& \metricci{0.190}{0.149}{0.234}
& \metricci{0.237}{0.190}{0.285}
& \metricci{0.278}{0.231}{0.329}
& \metricci{0.323}{0.272}{0.377}
& \secondmetricci{0.082}{0.064}{0.101}
& \metricci{0.718}{0.698}{0.738}
& \metricci{0.208}{0.185}{0.233} \\

Gaze prior only
& \metricci{0.158}{0.120}{0.199}
& \metricci{0.177}{0.136}{0.222}
& \metricci{0.209}{0.165}{0.253}
& \metricci{0.263}{0.215}{0.313}
& \metricci{0.068}{0.051}{0.088}
& \metricci{0.638}{0.615}{0.661}
& \metricci{0.163}{0.141}{0.186} \\

\midrule
Vote-MI~\cite{wang2024enhancing}
& \metricci{0.044}{0.022}{0.070}
& \metricci{0.190}{0.146}{0.234}
& \metricci{0.339}{0.288}{0.392}
& \metricci{0.465}{0.411}{0.519}
& \metricci{0.033}{0.027}{0.038}
& \metricci{0.472}{0.454}{0.491}
& \metricci{-0.025}{-0.036}{-0.015} \\

BioMedCLIP~\cite{zhang2025multimodal}
& \metricci{0.212}{0.168}{0.256}
& \metricci{0.282}{0.234}{0.332}
& \metricci{0.339}{0.288}{0.392}
& \metricci{0.443}{0.389}{0.500}
& \metricci{0.070}{0.058}{0.083}
& \metricci{0.658}{0.627}{0.688}
& \metricci{0.164}{0.142}{0.187} \\

3D U-Net~\cite{cciccek20163d}
& \secondmetricci{0.231}{0.187}{0.278}
& \metricci{0.275}{0.225}{0.326}
& \metricci{0.301}{0.250}{0.351}
& \metricci{0.361}{0.307}{0.415}
& \metricci{0.089}{0.071}{0.107}
& \bestmetricci{0.785}{0.763}{0.806}
& \bestmetricci{0.228}{0.204}{0.253} \\

3D Merlin + MLP~\cite{blankemeier2024merlin}
& \bestmetricci{0.247}{0.199}{0.294}
& \secondmetricci{0.297}{0.247}{0.348}
& \metricci{0.320}{0.269}{0.370}
& \metricci{0.399}{0.345}{0.453}
& \bestmetricci{0.094}{0.075}{0.114}
& \metricci{0.779}{0.755}{0.800}
& \secondmetricci{0.224}{0.201}{0.249} \\

Ours (w/o gaze cues)
& \metricci{0.199}{0.155}{0.244}
& \metricci{0.288}{0.237}{0.339}
& \secondmetricci{0.342}{0.288}{0.396}
& \metricci{0.468}{0.415}{0.525}
& \metricci{0.077}{0.065}{0.091}
& \metricci{0.768}{0.745}{0.791}
& \metricci{0.212}{0.189}{0.235} \\

Ours (w/ gaze cues)
& \metricci{0.228}{0.180}{0.275}
& \bestmetricci{0.339}{0.288}{0.392}
& \bestmetricci{0.421}{0.367}{0.475}
& \bestmetricci{0.563}{0.509}{0.617}
& \secondmetricci{0.091}{0.077}{0.106}
& \secondmetricci{0.781}{0.760}{0.801}
& \secondmetricci{0.224}{0.202}{0.246} \\

\midrule
\makecell[l]{\raisebox{-0.55em}{\textbf{$\Delta$ vs.}}\quad best 2D baseline}
& \posdelta{+1.6}
& \posdelta{+5.7}
& \sigposdelta{+7.9}
& \sigposdelta{+7.3}
& \posdelta{+0.9}
& \sigposdelta{+6.3}
& \sigposdelta{+1.5} \\[-0.1em]

\makecell[l]{\phantom{\textbf{$\Delta$ vs.}}\quad best 3D baseline}
& \negdelta{-1.9}
& \posdelta{+4.1}
& \sigposdelta{+10.1}
& \sigposdelta{+16.5}
& \negdelta{-0.3}
& \negdelta{-0.5}
& \negdelta{-0.5} \\
\bottomrule

\end{tabular*}
\end{table*}

\begin{figure*}[t!]
\centering

\begin{minipage}[t]{0.485\textwidth}
\centering
\includegraphics[width=\linewidth]{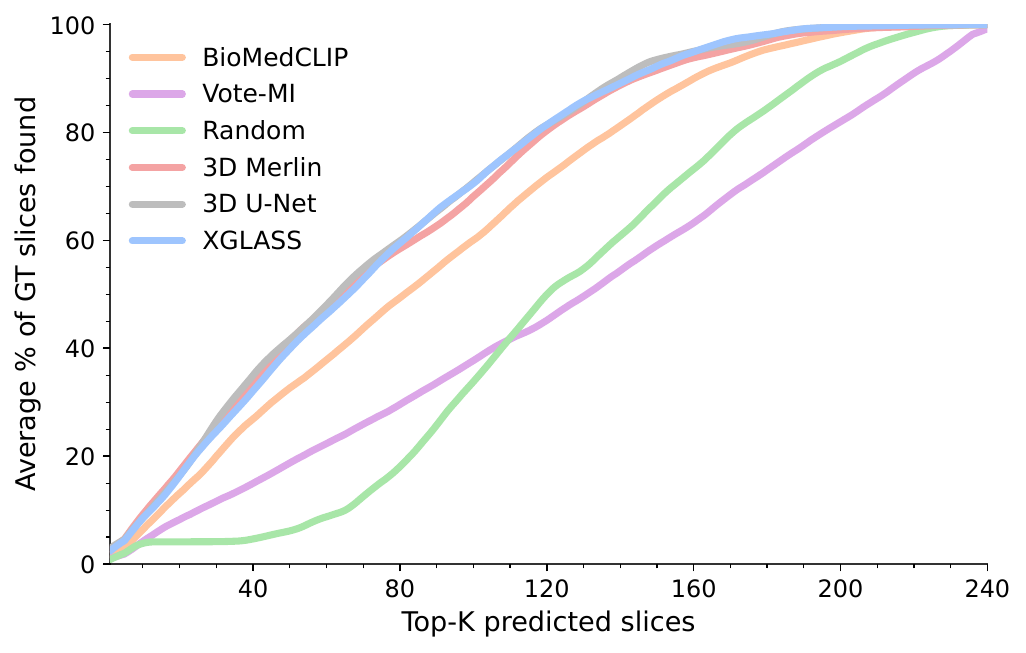}
\caption{
\textbf{Average Ground-Truth (GT) Coverage.}
For each CT, we compute the fraction of all GT slices recovered within the top-$K$ predictions. XGLASS outperforms the 2D baselines and compete with the 3D baselines despite substantially less compute.}
\label{fig:avg-gt-coverage}
\end{minipage}
\hfill
\begin{minipage}[t]{0.485\textwidth}
\centering
\includegraphics[width=\linewidth]{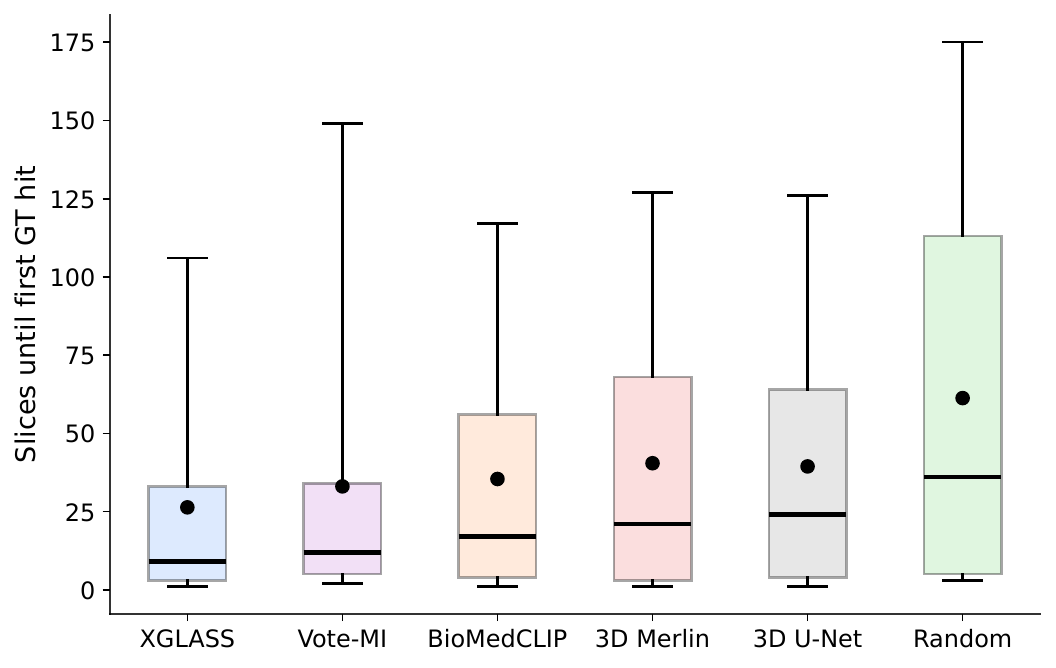}
\caption{
\textbf{Slices until first Ground-Truth (GT).}
Boxplots show the distribution of slices required to reach the first GT slice (median, mean, 25--75\% IQR, and 5--95\% whiskers). XGLASS reaches a relevant slice earlier and with lower variability than the baselines.}
\label{fig:first-hit-boxplot}
\end{minipage}

\end{figure*}

\subsection{2D-3D Cross-Modal Slice Sequencing}
Our model builds on a ViT-B/16 backbone with AugReg ImageNet-21k pretraining~\cite{dosovitskiy2020vit}, which processes the fused representation of the disease-conditioned importance map \(G\) and projected CT volume \(P\). To adapt the network from classification to slice-importance regression, we replace the original classification head with a lightweight two-layer MLP that outputs a fixed-length vector \(\hat{y}\in\mathbb{R}^{Z_{\text{target}}}\) (\(Z_{\text{target}}=240\)). A final sigmoid activation constrains all predictions to the range \([0,1]\), matching the normalized ground-truth slice–area profiles. The model receives a 3-channel \(224\times224\) image initialized with a coronal CT intensity projection \(P\), replicated across all three input channels. The corresponding disease-specific importance map \(G\) is then spatially scaled to the chest bounding-box region in the projection and blended into the red channel with an opacity coefficient \(\alpha\), while the green and blue channels remain unchanged. This channel-wise injection allows the ViT to process the anatomical projection and gaze-derived prior jointly in a single forward pass, enabling the model to estimate slice relevance directly from the fused representation. During training, the model obtains as input the fused \(G\) and \(P\) image, and learns to regress the ground-truth slice–area vector derived in Sec.~\ref{sec:data_prep_gt}. For each input image, the network outputs \(\hat{y}\in[0,1]^{240}\) and is optimized using MSE loss between \(\hat{y}\) and the corresponding normalized ground-truth vector \(y\). We train the full model end-to-end using the AdamW optimizer with a differential learning rate schedule, assigning a higher learning rate to the regression head (\(1\times10^{-4}\)) and a lower rate to the ViT backbone (\(1\times10^{-5}\)). Training is performed with a batch size of 16. This setup preserves the pretrained backbone’s representational capacity while allowing the head to adapt to the slice-importance regression task.

\section{Results}
\label{sec:Results}
We evaluate XGLASS on a held-out test set (split details in Supp. Sec.~\ref{sec:ct_dataset}) to answer the following questions: (i) How does XGLASS perform compared to existing 2D and 3D slice selection baselines such as Vote-MI~\cite{wang2024enhancing}, BioMedCLIP~\cite{zhang2025multimodal}, Merlin~\cite{blankemeier2024merlin}, 3D U-Net~\cite{cciccek20163d} and heuristic approaches (ii) does injecting 2D gaze–language priors improve 3D slice sequencing, and (iii) does this translate into practically useful retrieval of diagnostically relevant slices?

\begin{table*}[t!]
\centering
\caption{\textbf{Disease-Specific Results.} The ``Overall'' row corresponds to the ``Ours (w/ gaze cues)'' model from Table~\ref{tab:main_results}. Counts (N) are shown in parentheses. We report hit-rate metrics Hit@K, the averaged recall at 10 (AR@10), AUC and rank correlation (Spearman's $\rho$).}
\label{tab:category_results}
\begin{tabular*}{\textwidth}{@{\extracolsep{\fill}}lccccccc}
\toprule
\textbf{Category (N)} & \textbf{Hit@1} & \textbf{Hit@3} & \textbf{Hit@5} & \textbf{Hit@10} & \textbf{AR@10} & \textbf{AUC} & \textbf{Spear. ($\rho$)} \\
\midrule
Atelectasis (39)            & 0.205 & 0.385 & 0.436 & 0.564 & 0.075 & 0.773 & 0.265 \\
Consolidation (29)          & \underline{0.483} & \textbf{0.586} & \underline{0.655} & \textbf{0.759} & 0.114 & \textbf{0.810} & \textbf{0.386} \\
Groundglass opacity (95)    & 0.284 & 0.400 & 0.516 & \underline{0.674} & 0.114 & 0.728 & 0.272 \\
High lung volume (29)       & 0.172 & 0.241 & 0.276 & 0.414 & 0.055 & 0.657 & 0.161 \\
Lung nodule or mass (112)   & 0.107 & 0.205 & 0.286 & 0.446 & 0.074 & 0.722 & 0.129 \\
Pleural abnormality (12)    & \textbf{0.500} & \underline{0.583} & \textbf{0.667} & 0.667 & \textbf{0.139} & \underline{0.779} & \underline{0.345} \\
\midrule
Overall (316)               & 0.228 & 0.339 & 0.421 & 0.563 & 0.091 & 0.781 & 0.224 \\
\bottomrule
\end{tabular*}
\end{table*}

\subsection{Evaluation Metrics}
\label{sec:Metrics}
We used different metrics to assess how well XGLASS orders all axial slices and how useful this ordering is for retrieval in potential downstream tasks. Let $Z{=}240$ be the number of axial slices, $y\in[0,1]^Z$ the ground-truth slice-area vector, and $\hat{y}\in[0,1]^Z$ the predicted importance scores and a slice is \emph{positive} if $y_i>0$. All metrics are macro-averaged per study; AUC is computed over pooled slices. For overall ordering quality, we use ROC-AUC and Spearman’s rank correlation $\rho$~\cite{sedgwick2014spearman} between $y$ and $\hat{y}$ over all $Z$ slices, which directly evaluates agreement of the complete 240-slice ordering from most to least relevant. For practical analysis of performance on downstream tasks, we measure the hit rate at $K$ (Hit@$K$) by sorting slices in descending $\hat{y}$ and checking whether any positive slice appears among the first $K$ predicted slices. Further, we also calculate the average recall at $K$ (AR@$K$), indicating the fraction of all GT positive slices that are represented within the $K$ predictions. In summary, higher $\rho$ and ROC-AUC indicate better global ranking alignment, higher Hit@$K$ shows that a GT slice is in the $K$ predictions, and AR@$K$ reflects the percentage amount of all GT slices inside $K$ predictions.

\subsection{Baselines}
\label{sec:baselines}
We benchmark XGLASS against the state-of-the-art slice selection method Vote-MI~\cite{wang2024enhancing}, a BioMedCLIP-based slice-wise retrieval baseline~\cite{zhang2025multimodal}, heuristic and prior-only baselines, two 3D-native baselines~\cite{blankemeier2024merlin, cciccek20163d}, and a gaze-free variant of our model. Only the gaze-prior baseline and gaze-guided XGLASS use REFLACX gaze information; BioMedCLIP, 3D Merlin + MLP, 3D U-Net, and XGLASS are disease-conditioned, while XGLASS (w/out gaze), Vote-MI, Gaussian Centered, and Random are disease-agnostic. \textbf{Gaze prior only} ranks slices using only the disease-specific gaze-derived superior--inferior profile, without patient-specific CT input, serving as a control for static anatomical prior effects. \textbf{Vote-MI} was originally proposed for brain MRI; we follow its preprocessing and two-stage slice-selection procedure, but adapt it to thoracic CT by replacing the original feature extractor with MedVAE~\cite{varma2025medvaeefficientautomatedinterpretation}, a medical foundation model, and convert the selected representatives into continuous slice scores for fair metric-based comparison. \textbf{BioMedCLIP} is used as a slice-wise image--text retrieval baseline. For each CT-condition pair, we encode all axial slices with the BioMedCLIP image encoder and the queried disease text with the text encoder, and then rank slices by image--text cosine similarity to obtain a disease-conditioned 240-dimensional relevance score vector. Full prompt and implementation details are provided in Supplementary Sec.~\ref{sec:biomedclip_details}. For the 3D-native model comparisons, \textbf{3D Merlin + MLP} uses a frozen Merlin volumetric encoder~\cite{blankemeier2024merlin} followed by a disease-conditioned MLP trained to regress the same 240-dimensional slice-relevance target as XGLASS, and \textbf{3D U-Net}~\cite{cciccek20163d} uses a 3D U-Net encoder trained from scratch on ReXGroundingCT with a slice-regression head for the same targets. Further implementation details for both 3D baselines are provided in Supplementary Sec.~\ref{sec:3d_baseline_details}. \textbf{XGLASS (w/o gaze cues)} uses the identical backbone, projections, loss, and training as XGLASS but disables the gaze blend ($\alpha{=}0$), isolating the architecture without disease-specific importance priors. \textbf{Gaussian Centered} places a Gaussian bump over slice indices with fractional width $\sigma_{\text{frac}}$, yielding higher scores near the mid-thorax, and \textbf{Random} draws an independent score for each slice from $\mathrm{Uniform}[0,1]$ to reflect chance performance. All methods output a 240-dimensional score vector per study and are evaluated with the same metrics (ROC-AUC, Spearman's $\rho$, Hit@K, and AR@K), as described in Section~\ref{sec:Metrics}.

\begin{figure*}[t]
  \centering
  \includegraphics[width=\linewidth]{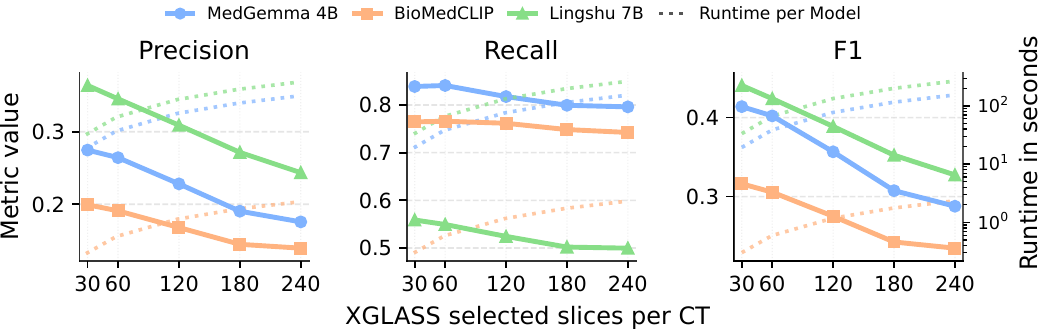}
\caption{
\textbf{Downstream slice-wise VLM screening under XGLASS budgets.}
For each CT-condition pair, XGLASS selects the Top-$K$ axial slices and the downstream model is evaluated only on this subset. Precision, recall, F1, and runtime are reported for MedGemma 4B, Lingshu 7B, and BioMedCLIP under the same single-GPU setting. Smaller XGLASS budgets consistently improve precision and F1, while reducing runtime by decreasing the number of inspected slices. Comparison to other slice selectors are in Supplementary Sec.~\ref{sec:matched_budget_downstream}.
}
  \label{fig:downstream_screening_metrics}
\end{figure*}

\subsection{Overall Performance}
Table~\ref{tab:main_results} shows that gaze-guided XGLASS outperforms the 2D baselines across all reported metrics, with paired bootstrap intervals indicating significant gains in Hit@5, Hit@10, AUC, and Spearman's $\rho$. The native 3D baselines are competitive on global ranking metrics, with 3D U-Net achieving the highest AUC and Spearman's $\rho$ point estimates and 3D Merlin achieving the highest Hit@1 and AR@10 point estimates. Despite operating on the full CT volume, the 3D baselines do not translate their small non-significant advantages in Hit@1 and global ranking into reliable slice retrieval and XGLASS remains significantly better on Hit@5 and Hit@10, the most relevant metrics for prioritizing a compact inspection set. Figure~\ref{fig:first-hit-boxplot} further shows that XGLASS reaches the first diagnostically relevant slice earlier and more consistently than all baselines. In the coverage analysis (Figure~\ref{fig:avg-gt-coverage}), the 3D baselines become competitive, but together with their later first-hit distributions in Figure~\ref{fig:first-hit-boxplot}, it suggests that the 3D models recover dense relevant regions once the correct location is found, but are less reliable at finding that location in every case. A case-level Miss@10 analysis supports this interpretation: Merlin and 3D U-Net miss the early inspection budget in 190/316 and 202/316 cases, respectively, compared with only 138/316 missed cases for XGLASS, indicating that their slightly higher Hit@1, AR@10, and global-ranking point estimates do not translate into more reliable compact-budget retrieval. For clinical use cases, reliably surfacing at least one diagnostic slice within the inspected budget is more valuable than dense coverage only after successful localization, making XGLASS more suitable for practical slice selection, especially since it also requires 34.8$\times$ less inference time and 6.5$\times$ less GPU memory than Merlin, and 130.8$\times$ less time and 10.9$\times$ less memory than U-Net.

\subsubsection{Disease-specific Performance}
The per-category trends in Table~\ref{tab:category_results} align with typical anatomical distributions. \emph{Consolidation and Pleural abnormalities} are often gravity-dependent and cluster in posterior lower lobes, producing a relatively narrow and predictable band along the $z$-axis. This yields the highest AUC/$\rho$ and strong Hit@K performance. \emph{Ground-glass opacities} are frequently diffuse and extend across adjacent slices, which lifts mid- to higher-$K$ retrieval (e.g., Hit@10) but softens global ranking metrics. In contrast, \emph{lung nodules/masses} are small, sparse, and spatially unpredictable along the $z$-axis, which reduces the chance that the very top-ranked slices contain a hit and depresses global rank agreement. Thus, lung nodules are relatively less suited to our approach due to their small size and variable location, characteristics that also make them challenging for slice-selection methods in general. When evaluated only on the lung-nodule subset, XGLASS achieves the strongest Hit@K and AUC, showing that despite performing worse than on other diseases, it still provides the best slice prioritization (Supplementary Sec.~\ref{sec:nodule_baselines}). Taken together, these findings indicate that XGLASS learns disease-specific $z$-axis priors similar to radiologist behaviour, prioritizes anatomically informative regions, and therefore performs best for conditions with consistent spatial distributions such as consolidations.

\subsubsection{Computational Efficiency}
\label{sec:efficiency}
XGLASS contains approximately 86M parameters and is trained on a single 2D projection per CT study. On an NVIDIA RTX 4080, one training epoch over 2,512 projections with batch size 16 takes 62.2\,s, corresponding to roughly 40 samples/s, with approximately 3\,GB peak GPU memory. Early stopping on validation loss consistently selects checkpoints after 6--8 epochs, yielding a complete training run in under 10 minutes. At inference time, XGLASS processes a full CT study, including intensity projection, in approximately 0.025\,s per finding (0.150\,s for all findings) with an estimated cost of 33.7\,GFLOPs and 0.34\,GB memory. In comparison, Vote-MI requires 7.51\,s and approximately 116.33\,TFLOPs per CT, while slice-wise BioMedCLIP requires 1.40\,s and approximately 8.09\,TFLOPs per CT. The 3D Merlin and U-Net baselines require 0.87\,s and 3.27\,s per finding, respectively, with 2.21\,GB and 3.72\,GB peak inference memory, making them 34.8$\times$/130.8$\times$ slower and 6.5$\times$/10.9$\times$ more memory-intensive than XGLASS. Overall, XGLASS is orders of magnitude cheaper than slice-wise 2D baselines and substantially faster and more memory-efficient than the 3D baselines, making local retraining on commodity GPUs feasible for institutions with private clinical data.

\subsubsection{Downstream Slice-wise VLM Screening}
\label{sec:downstream_vlm_screening}
We evaluate XGLASS as a disease-conditioned pre-filter for slice-wise VLM screening with 2D VLMs. For each CT-condition pair, XGLASS ranks all axial slices by relevance to the queried finding, and the downstream model is evaluated only on the Top-$K$ selected slices. Metrics are computed over the inspected slices and compared against applying the same model to all 240 slices. Prompts, detailed inference settings and a matched-budget comparison to other slice selection methods are provided in Supplementary Sec.~\ref{sec:baseline_downstream_details}. As shown in Fig.~\ref{fig:downstream_screening_metrics}, smaller XGLASS budgets consistently improve precision across MedGemma 4B~\cite{sellergren2025medgemma}, Lingshu 7B~\cite{xu2025lingshu}, and BioMedCLIP~\cite{zhang2025multimodal}, indicating that XGLASS determines disease-enriched slice subsets. This also improves F1 over inspected slices at small budgets: Top-30 improves F1 from 0.288 to 0.414 for MedGemma, from 0.327 to 0.441 for Lingshu, and from 0.235 to 0.316 for BioMedCLIP. Compared with other slice-selected subsets, XGLASS improves average Top-30 F1 over the strongest 2D selector from 0.338 to 0.390, while remaining competitive to the strongest 3D selector. Evaluating fewer slices also reduces downstream runtime, with Top-30 requiring only 12.5\% of the slice-wise model calls compared with all-slice inference. This supports XGLASS as a fast pre-selection step that enriches disease-relevant slices before costly slice-wise VLM review, 2D dataset curation or similar applications.

\subsection{Ablation Studies}
We perform targeted ablations to isolate the effect of the main XGLASS design choices. Hyperparameter choices were made using the validation split resulting in ViT-B/16 backbone, AIP-wide CT projection, gaze blending strength of $\alpha{=}0.75$, and training using MSE. The corresponding validation results and selection criteria are reported in Supplementary Sec.~\ref{sec:supp_validation_selection}. All ablations tables modify only one factor at a time, and are reported on the held-out test set. Additional backbone, multi-channel-fusion, gaze-prior perturbations, loss ablation, and gaze prior-specificity ablations are provided in Supplementary Sec.~\ref{sec:extendedab}.

\subsubsection{Weak Supervision.}
\label{sec:weak_supervision}
\noindent Our best training target uses segmentation-derived slice-area scores, but XGLASS can also be trained from substantially weaker annotations. We evaluate three reduced-supervision targets: \emph{SE-Binary}, which marks all slices between annotated start and end positions as positive; \emph{SE-Smooth}, which additionally applies Gaussian smoothing to this interval; and \emph{Rel-Rank}, which uses relative slice ranking instead of dense area scores. As shown in Table~\ref{tab:weak_supervision}, reduced-supervision targets remain competitive, highlighting XGLASS's ability to be trained from weak annotations.

\begin{table}[h]
\centering
\caption{\textbf{Supervision Target Ablation.} Comparison of dense Slice-Area supervision with only start/end slice annotations.}
\label{tab:weak_supervision}
\setlength{\tabcolsep}{3pt}
\renewcommand{\arraystretch}{1.05}
\small
\resizebox{\columnwidth}{!}{%
\begin{tabular}{lccccc}
\toprule
\textbf{Target} & \textbf{AUC} & \textbf{Spear.} & \textbf{Hit@1} & \textbf{Hit@5} & \textbf{Hit@10} \\
\midrule
SE-Binary & 0.791 & 0.223 & 0.168 & 0.364 & 0.490 \\
SE-Smooth & \underline{0.792} & \textbf{0.227} & \underline{0.220} & 0.413 & \underline{0.524} \\
Rel-Rank & \textbf{0.793} & 0.212 & 0.187 & \underline{0.417} & \underline{0.524} \\
Slice-Area & 0.781 & \underline{0.224} & \textbf{0.228} & \textbf{0.421} & \textbf{0.563} \\
\bottomrule
\end{tabular}}
\end{table}

\subsubsection{Volume Projection Ablation}
\label{sec:CTPablations}
The results in Table~\ref{tab:ablation_projection} show a trade-off between global ranking and early retrieval utility. \textbf{Beer–Lambert} achieves the highest AUC (0.795) and Spearman’s $\rho$ (0.232), and it edges out others at Hit@5 (0.427), indicating strong overall separability and mid-$K$ retrieval. In contrast, \textbf{AIP (wide)} delivers the best \emph{utility-centric} performance with the top Hit@1 (0.228), Hit@10 (0.563), and near-best Hit@5 (0.421), suggesting that preserving a broad intensity range helps rank diagnostically relevant slices.
\begin{table}[h]
  \centering
  \caption{\textbf{CT Projection Ablation.} Comparison of Beer--Lambert and AIP projections with lung, soft-tissue, and wide CT window.}
  \label{tab:ablation_projection}
  \setlength{\tabcolsep}{2pt}
  \renewcommand{\arraystretch}{1.05}
  \small
  \resizebox{\columnwidth}{!}{%
  \begin{tabular}{lcccccc}
    \toprule
    \textbf{Projection} & \textbf{AUC} & \textbf{Spear.} & \textbf{Hit@1} & \textbf{Hit@5} & \textbf{Hit@10} \\
    \midrule
    AIP (lung)   & \underline{0.784} & 0.222 & 0.206 & 0.361 & 0.462  \\
    Beer-Lambert     & \textbf{0.795}    & \textbf{0.232} & 0.196 & \textbf{0.427} & \underline{0.535}  \\
    AIP (soft tissue)   & 0.783  & 0.223 & \underline{0.228} & 0.383 & 0.519  \\
    \textbf{AIP (wide) } & 0.781       & \underline{0.224} & \textbf{0.228} & \underline{0.421} & \textbf{0.563} \\
    \bottomrule
  \end{tabular}
  }
\end{table}

\subsubsection{Gaze Fusion Ablation}
\label{sec:gaze_fusion_ablation} As described in Section~\ref{sec:heatmap_gen}, XGLASS fuses the disease-specific gaze prior with the coronal CT projection by alpha-blending the importance map into one RGB input channel. Table~\ref{tab:ablation_gaze_fusion} evaluates both the blending strength and the injection channel. Adding gaze guidance consistently improves over the gaze-free setting ($\alpha{=}0$), confirming that the disease-conditioned prior provides useful localization information. While $\alpha{=}1.0$ achieves the highest global AUC and Spearman's $\rho$, but it partially suppresses CT intensity cues lowering retrieval metrics. In contrast, $\alpha{=}0.75$ gives the best early retrieval performance, reaching Hit@10 of 0.563. Among single-channel injection variants, red-channel fusion performs best, outperforming green and blue injection on all metrics. Additional multi-RGB-channel fusion results are reported in the Supplementary Material in Sec.~\ref{sec:ablation_channel_injection}.

\begin{table}[h]
  \centering
\caption{\textbf{Gaze fusion ablation.} Effect of blending opacity $\alpha$ and RGB injection channel for the disease-specific importance map.}
  \label{tab:ablation_gaze_fusion}
  \setlength{\tabcolsep}{2pt}
  \renewcommand{\arraystretch}{1.05}
  \small
  \resizebox{\columnwidth}{!}{%
  \begin{tabular}{lcccccc}
    \toprule
    \textbf{Channel} & $\boldsymbol{\alpha}$ & \textbf{AUC} & \textbf{Spear.} & \textbf{Hit@1} & \textbf{Hit@5} & \textbf{Hit@10} \\
    \midrule
       & 0.00 & 0.768 & 0.212 & \underline{0.199} & 0.342 & 0.468 \\
       & 0.25 & \underline{0.787} & 0.221 & 0.187 & 0.392 & 0.506 \\
    Red   & 0.50 & 0.785 & 0.220 & 0.196 & \underline{0.418} & 0.509 \\
       & 1.00 & \textbf{0.797} & \textbf{0.227} & \underline{0.199} & 0.415 & \underline{0.528} \\
       & 0.75 & 0.781 & \underline{0.224} & \textbf{0.228} & \textbf{0.421} & \textbf{0.563} \\
    \midrule
    Green & 0.75 & 0.754 & 0.212 & 0.193 & 0.345 & 0.459 \\
    Blue  & 0.75 & 0.761 & 0.209 & 0.190 & \underline{0.418} & 0.519 \\
    \bottomrule
  \end{tabular}
  }
\end{table}

\section{Conclusion and Limitations}
\label{sec:Conclusion}

We presented XGLASS, a lightweight disease-conditioned model that transfers aggregated CXR gaze-language priors to CT for relevance-ordered axial slice retrieval. XGLASS significantly improves Hit@5 and Hit@10 over the strongest evaluated 2D and 3D baselines while remaining computationally efficient. Downstream experiments further show that XGLASS enriches compact disease-relevant candidate sets, improves downstream classification performance, and reduces slice-wise model calls. Beyond slice selection, XGLASS illustrates how heterogeneous 2D supervision can help bridge the gap between 2D foundation models and 3D medical image analysis. Nevertheless, several limitations remain. Despite surveying publicly available CT datasets, ReXGroundingCT is the only dataset, that overlaps with the 14 REFLACX-derived disease priors, yielding only to one dataset and six shared thoracic findings for evaluation. Our downstream screening experiments provide an initial demonstration of XGLASS utility and not direct use cases; broader applications such as VQA, report generation, retrieval, and localization remain future work.

\section*{Acknowledgments}
This work was funded by the Deutsche Forschungsgemeinschaft
(DFG, German Research Foundation) -- GRK 3012 -- 520750254
(\url{https://kemai.uni-ulm.de/}).

\fi

{
    \small
    \bibliographystyle{ieeenat_fullname}
    \bibliography{main}
}
\clearpage
\maketitlesupplementary

\appendix
\setcounter{section}{0}

\begin{figure*}[ht]
    \centering
    \includegraphics[width=\linewidth]{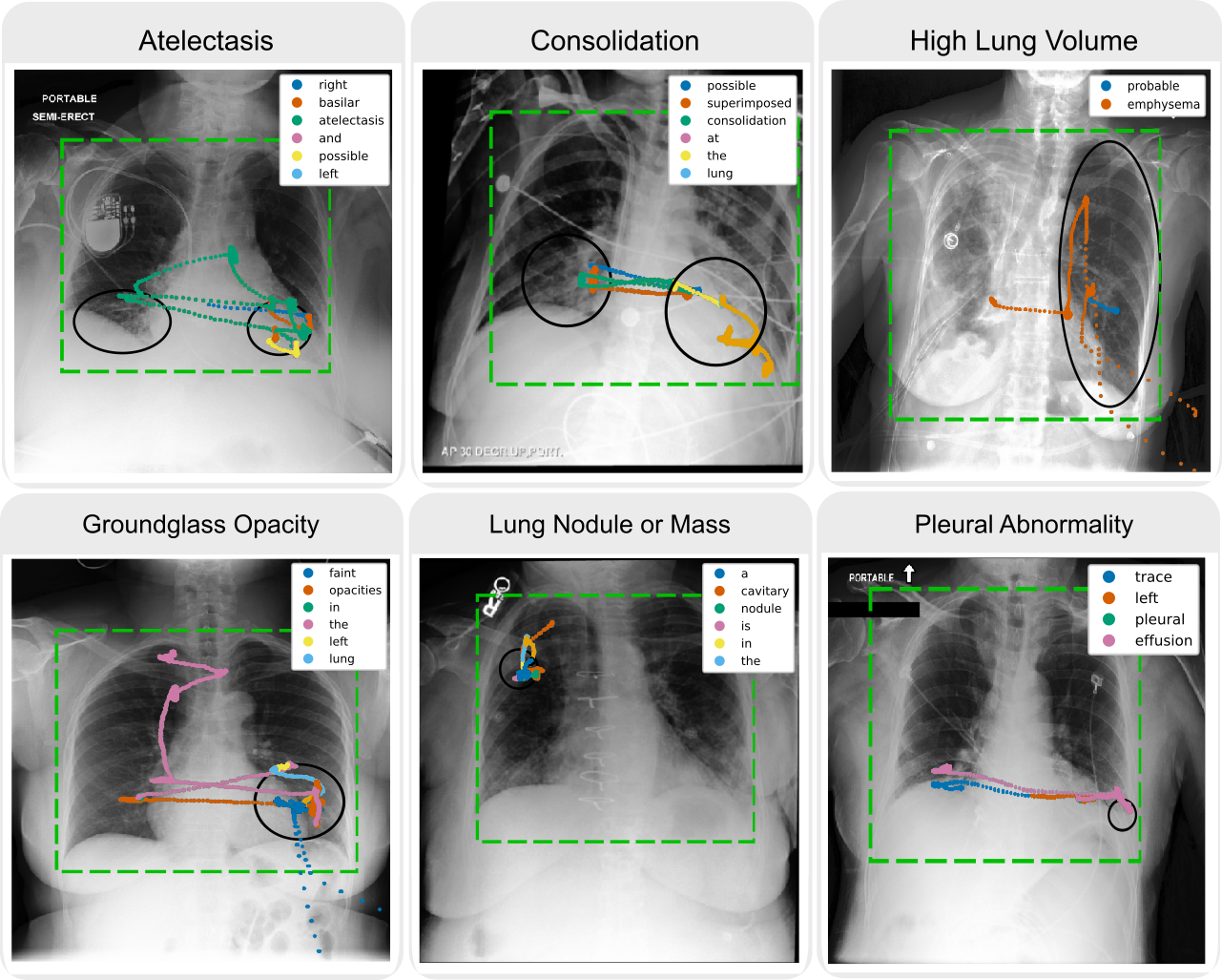}
     \caption{
    \textbf{Text–gaze alignment for disease phrases.}
    Examples of text–gaze alignment used to construct disease-conditioned importance maps for the six target findings: Atelectasis, Consolidation, Groundglass opacity, Lung nodule or mass, High lung volume, and Pleural abnormality.
    Each panel shows a REFLACX chest X-ray with the chest bounding box (green dashed line), disease-specific annotation ellipses (black), and gaze trajectories for the words in the corresponding disease phrase (color-coded by word with the pre-speech buffer in orange).
    Fixations that fall outside the chest bounding box, visible for the High lung volume and Groundglass opacity examples, are removed during preprocessing as described in Section~3.1.
    }
    \label{fig:text_gaze_heatmaps}
\end{figure*}

\section{Qualitative Results}
\begin{figure*}[t]
  \centering
  \includegraphics[width=\linewidth]{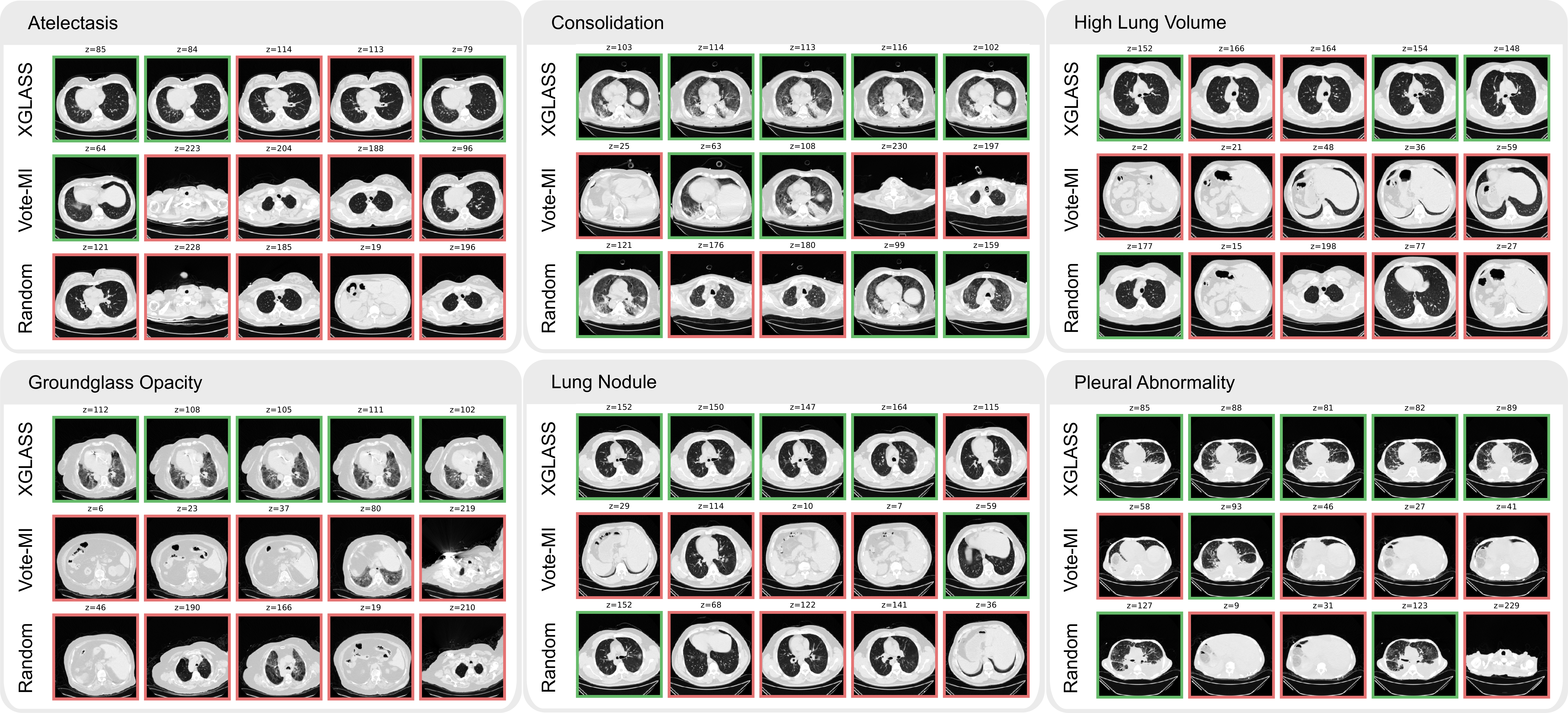}
  \caption{
  \textbf{Qualitative Results.} Slice retrieval across six chest CT findings: Atelectasis, Consolidation, Ground-glass Opacity, Lung Nodule/Mass, High Lung Volume, and Pleural Abnormality. For each category, we show the axial slice indices ($z$) predicted by a Random baseline, Vote-MI, and our XGLASS selector. XGLASS not only identifies a ground-truth slice (green border) earlier, but also provides broader coverage of the ground-truth region by retrieving multiple slices that fall within it, rather than a single correct slice.}
  \label{fig:qualitative}
\end{figure*}

We complement the quantitative metrics with a qualitative analysis of slice selection across six chest CT findings. Figure~\ref{fig:qualitative} visualizes the axial slice indices ($z$) selected by a Random baseline, Vote-MI, and XGLASS. Across categories, XGLASS not only hits a ground-truth slice earlier, but also concentrates its selections within the annotated region, yielding denser and more contiguous coverage than competing methods. These examples illustrate that our model prioritizes clinically meaningful slices rather than scattered or isolated hits, aligning with the intended notion of region-level localization.

\section{Disease-Conditioned Importance Maps}
\label{sec:reflacx_processing}

\subsection{Text--Gaze Alignment}

To construct disease-conditioned 2D importance maps from REFLACX~\cite{bigolin2022reflacx}, we jointly process the word-level transcripts and synchronized gaze recordings. For each target disease, we first detect positive (non-negated) mentions in the transcript using label-specific regular expressions together with clause-level negation handling, as described in Section~\ref{sec:heatmap_gen}. Each detected mention defines a speech window $[t_{\text{start}} - \tau_{\text{pre}},\, t_{\text{end}}]$ with a short pre-speech buffer $\tau_{\text{pre}}$ of 2 seconds to account for anticipatory gaze. We then collect all gaze samples whose timestamps fall inside this window and project their screen coordinates onto the corresponding chest X-ray. Following the REFLACX protocol, we normalize coordinates to the chest bounding box and discard samples that fall outside this region (e.g., blink artefacts or fixations off the image). Figure~\ref{fig:text_gaze_heatmaps} shows qualitative examples of the resulting text–gaze alignment for the six disease categories used in this work.

\subsection{Extracted Gaze Importance Maps}
Using label-specific regular expressions and clause-level negation on the REFLACX reports, we defined 14 non–device pathology categories and, for each of them, extracted a disease-specific importance map by aggregating gaze samples associated with positive mentions across all matching sessions.
Figure~\ref{fig:reflacx_gaze_counts} summarizes how many gaze–report pairs contribute to each category.

\subsection{AP--PA Robustness of CXR Gaze Priors}
\label{sec:supp_ap_pa}

\begin{figure*}[t]
    \centering
    \includegraphics[width=\linewidth]{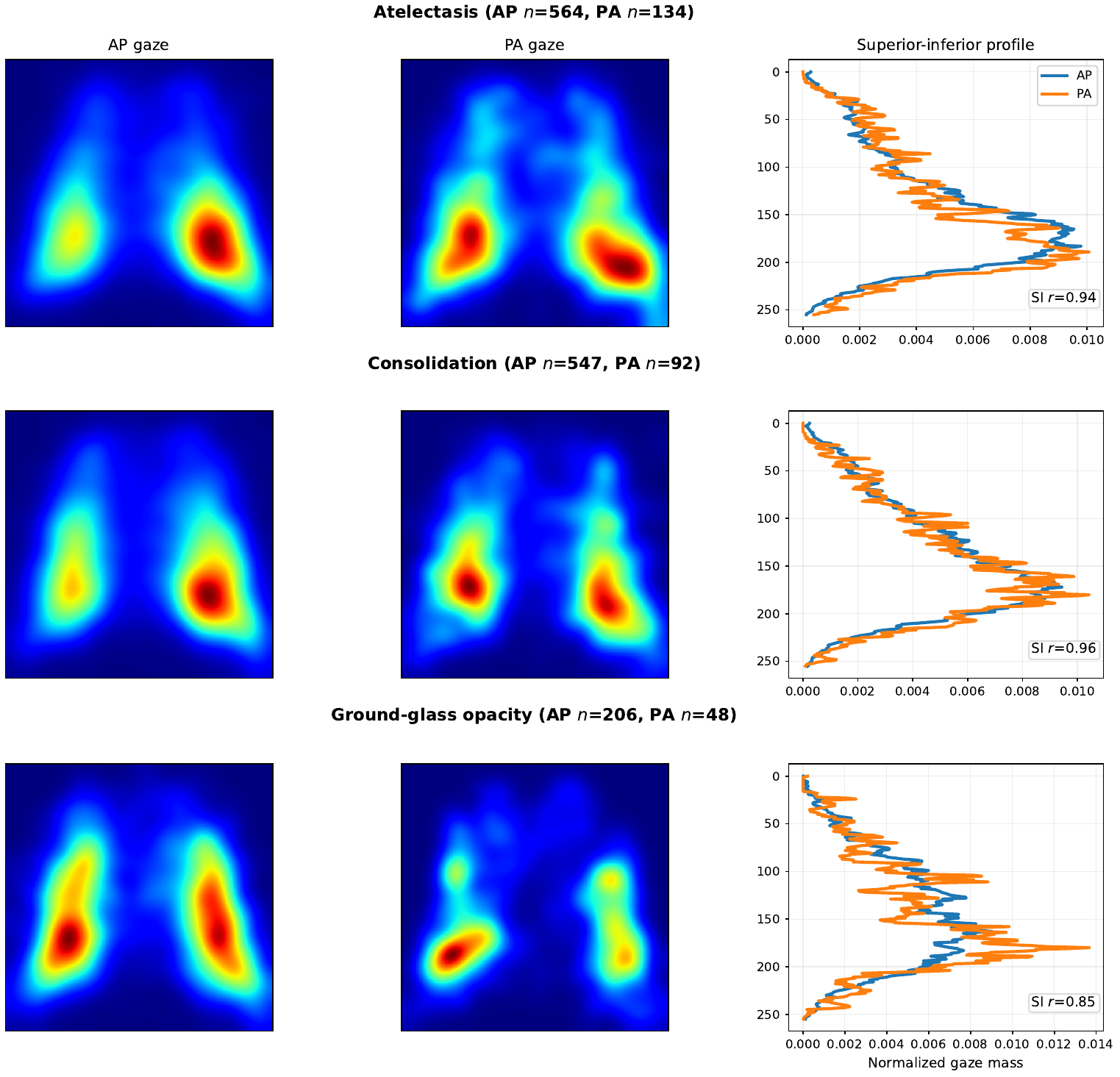}
\caption{
\textbf{AP--PA comparison of disease-conditioned gaze priors, part 1.}
For each finding, we construct separate AP-only and PA-only REFLACX gaze heatmaps and compare their superior--inferior profiles. The superior--inferior axis encodes the axial slices and therefore it is most relevant for XGLASS.
}
    \label{fig:ap_pa_heatmaps_1}
\end{figure*}

\begin{figure*}[t]
    \centering
    \includegraphics[width=\linewidth]{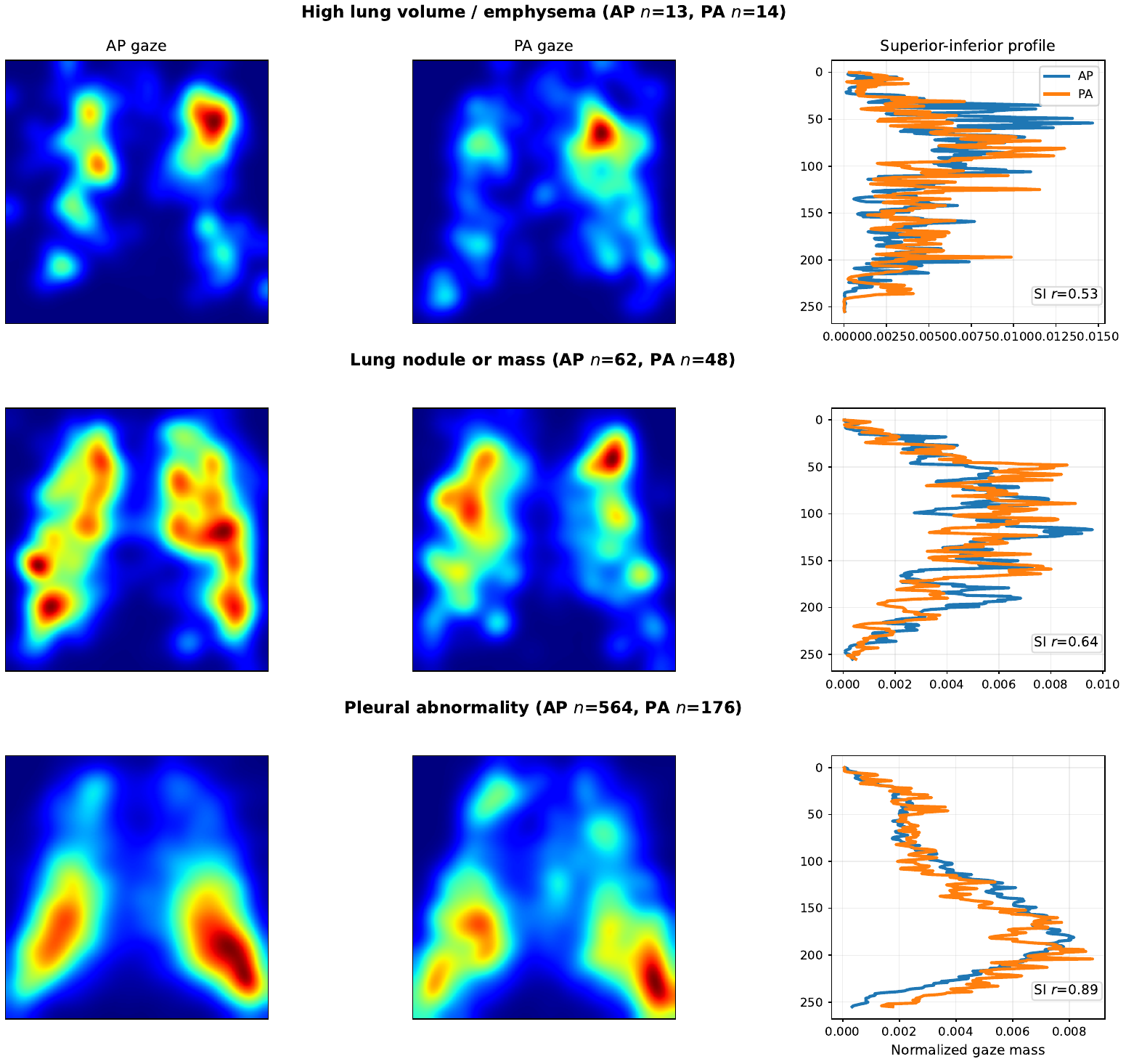}
\caption{
\textbf{AP--PA comparison of disease-conditioned gaze priors, part 2.}
Continuation of the AP--PA comparison for the remaining findings. Similar superior--inferior profiles across views support combining AP and PA sessions into more stable disease-conditioned gaze priors.
}
    \label{fig:ap_pa_heatmaps_2}
\end{figure*}

Because REFLACX is derived from frontal chest radiographs, one potential concern is that the resulting gaze priors may depend on acquisition view. AP and PA radiographs can differ in apparent anatomy due to projection geometry, magnification, and patient positioning. We therefore stratify REFLACX sessions by MIMIC-CXR view position and construct separate AP-only and PA-only disease-conditioned gaze maps. Figures~\ref{fig:ap_pa_heatmaps_1} and~\ref{fig:ap_pa_heatmaps_2} compare the resulting maps for all six target findings. Despite view-dependent differences in the 2D heatmaps, the superior--inferior profiles remain similar across AP and PA views, with correlations ranging from $r{=}0.53$ to $0.96$. Since this axis is the one used for axial slice ordering, these results support combining AP and PA sessions to obtain more stable disease-conditioned gaze priors for XGLASS.

\subsection{CXR--CT Gaze Alignment}
\label{sec:supp_cxr_ct_gaze_alignment}

\begin{figure*}[t]
    \centering
    \includegraphics[width=\linewidth]{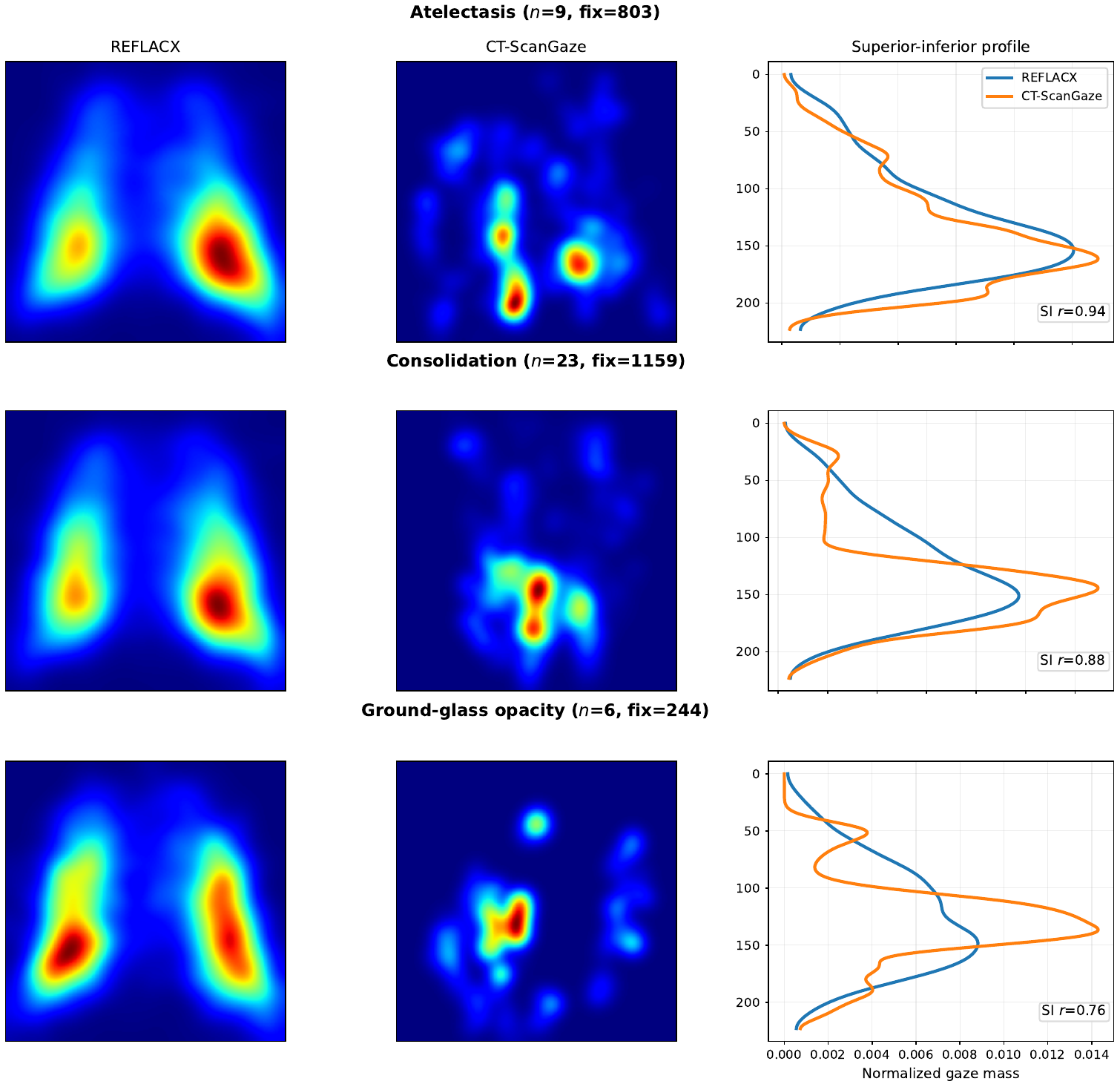}
\caption{
\textbf{CXR--CT comparison of disease-conditioned gaze priors, part 1.}
We compare REFLACX CXR-derived gaze priors with sentence-matched CT-ScanGaze fixation heatmaps. Although CT reading follows different fixation dynamics, the superior--inferior profiles show similar coarse disease-specific attention patterns, supporting the idea to transfer CXR gaze priors to CTs.
}
    \label{fig:ct_scangaze_reflacx}
\end{figure*}

\begin{figure*}[t]
    \centering
    \includegraphics[width=\linewidth]{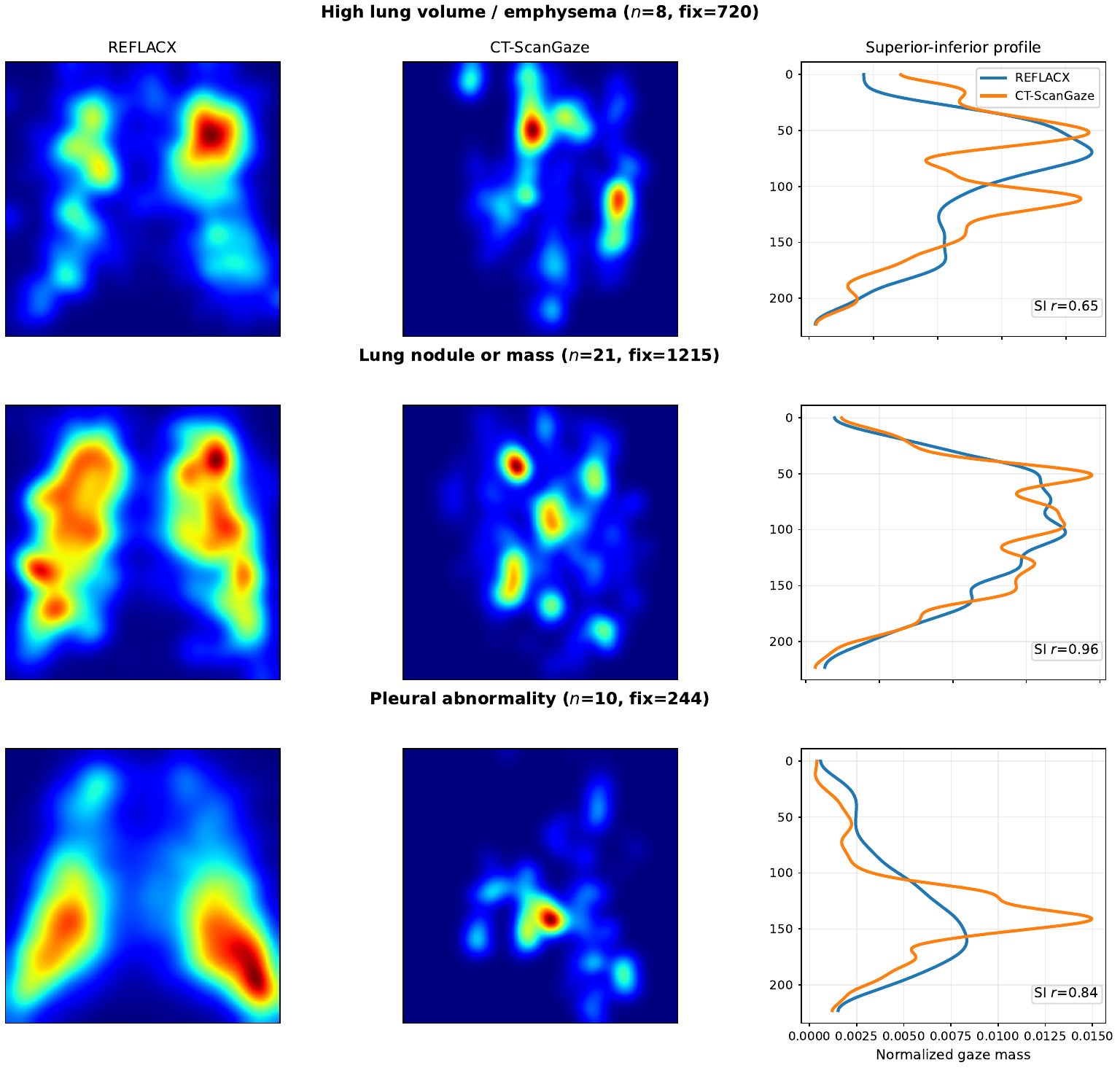}
\caption{
\textbf{CXR--CT comparison of disease-conditioned gaze priors, part 2.}
Continuation of the CXR--CT comparison for the remaining findings. The aligned superior--inferior profiles support transferring CXR-derived disease-specific search priors to CT slice selection.
}
    \label{fig:ct_scangaze_reflacx_2}
\end{figure*}

To further assess whether CXR-derived gaze priors capture transferable anatomical search patterns, we compare them with CT-ScanGaze~\cite{pham2025ct}, a CT eye-tracking dataset with timestamped dictation. CT-ScanGaze contains 57 unique CT cases, but several scans have inconsistent orientation metadata or cover non-chest anatomy. We therefore performed visual quality control and excluded cases with incorrect projection geometry or non-thoracic coverage. For the remaining cases, we constructed disease-conditioned CT gaze maps analogously to the REFLACX priors: report sentences were matched to gaze by timestamp, using the same 2\,s pre-speech buffer, and fixations were aggregated for the corresponding finding. These CT-ScanGaze maps provide insight into CT gaze patterns but are substantially sparser than REFLACX, with 6--23 CT cases and 244--1{,}215 fixations per finding (mean: 12.8 cases and 731 fixations). Figures~\ref{fig:ct_scangaze_reflacx}--\ref{fig:ct_scangaze_reflacx_2} compare the resulting CT-ScanGaze maps with the REFLACX CXR priors. The 2D heatmaps differ visibly, reflecting different reading behavior in CT: radiologists often fixate a local region and scroll through adjacent axial slices, rather than scanning broadly across the frontal image as in CXR. However, the superior--inferior gaze-density profiles remain aligned across modalities, with correlations ranging from 0.65 to 0.96 for the analyzed findings. This indicates that although CT and CXR differ in traversal order and fixation dynamics, radiologists still attend to similar disease-relevant anatomical regions. Thus, the CXR-derived gaze priors encode non-arbitrary, disease-specific search patterns that can be transferred to CT and used as additional input for axial slice selection.

\begin{figure}[t]
    \centering
    \includegraphics[width=\linewidth]{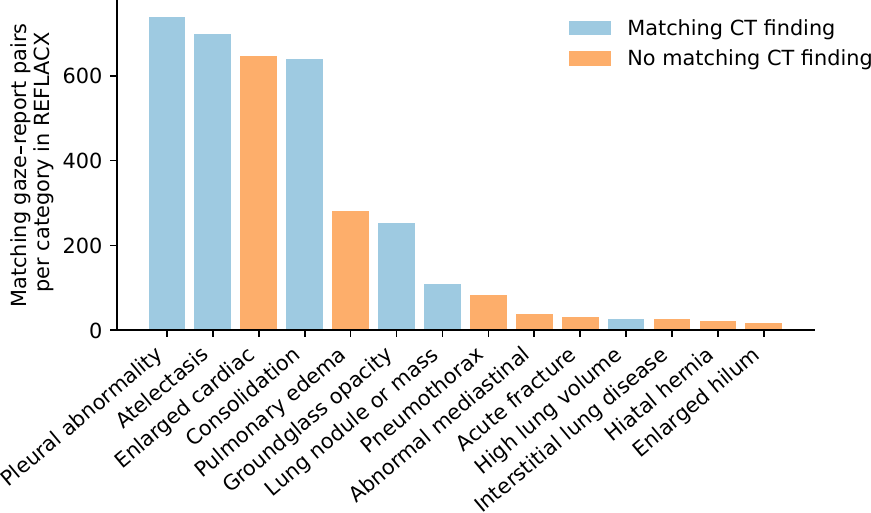}
    \caption{
    \textbf{REFLACX sessions per disease category.}
    Number of REFLACX sessions contributing to each disease-conditioned gaze heatmap.
    Bars show the number of matching gaze–report pairs for each category defined in REFLACX.
    Categories highlighted in blue correspond to the six findings used in XGLASS, whereas orange bars denote additional categories for which we obtained gaze heatmaps but did not use them due to the lack of corresponding findings in the CT dataset.
    }
    \label{fig:reflacx_gaze_counts}
\end{figure}

\label{sec:ct_dataset_projections}

\begin{figure}[t]
  \centering
  \includegraphics[width=\linewidth]{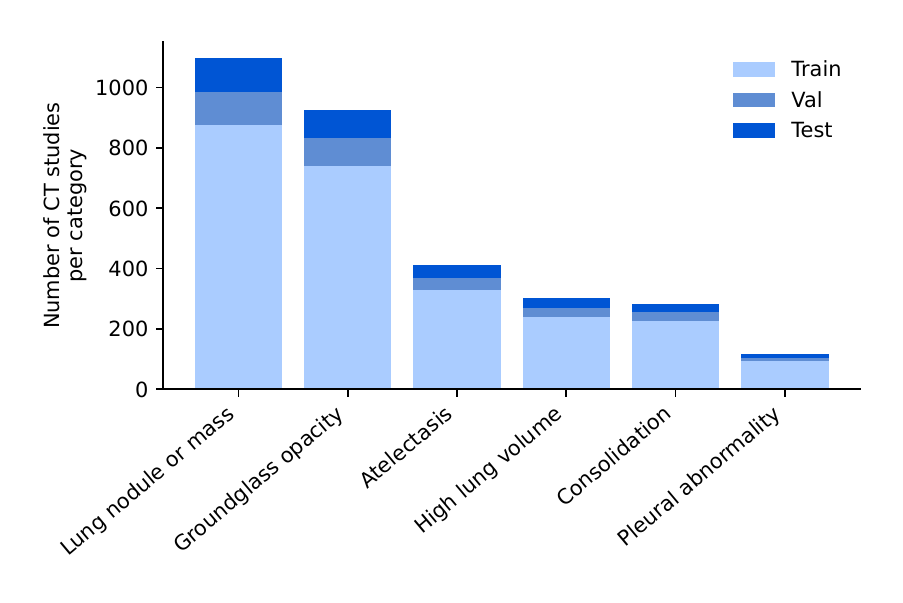}
\caption{
\textbf{CT disease category distribution.}
Counts of CT--finding cases for the six ReXGroundingCT categories selected for XGLASS, stacked by train/validation/test split. These categories were retained because they overlap with the REFLACX-derived gaze priors and provide sufficient CT cases for training and evaluation.
}
  \label{fig:cat_dist}
\end{figure}

\section{CT Dataset and Volume Projections}
\label{sec:supp_ct_dataset}

\subsection{Dataset Selection}
To evaluate how effectively our disease-specific eye-tracking priors transfer from CXR to CT, we require a CT dataset that (i) provides report text and findings aligned with individual volumes, and (ii) contains findings that can be reliably matched to our REFLACX-derived disease-specific importance maps. As discussed in Section~\ref{sec:volumeprojection}, ReXGroundingCT~\cite{baharoon2025rexgroundingct} is, to our knowledge, the only large-scale CT dataset that satisfies both criteria. Because this dataset will also be used to train XGLASS, each disease category must be supported by a sufficient number of CT studies. We therefore exclude any finding category with fewer than 100 matching CTs. As shown in Figure~\ref{fig:reflacx_gaze_counts} and Figure~\ref{fig:cat_dist}, only six of the 14 REFLACX-defined categories reach this threshold in our standardized 240-slice ReXGroundingCT subset introduced in Section~\ref{sec:volumeprojection}, and we consequently restrict our experiments to these six categories.\\

\subsection{Training Dataset and Splits}
\label{sec:ct_dataset}
We use an 80/10/10 CT-volume-level train--validation--test split of the standardized 240-slice ReXGroundingCT subset, with no volume overlap between partitions and all findings from each CT retained in the same split. This yields 2{,}514/306/316 CT--finding cases for training/validation/testing, respectively, totaling 3{,}136 cases across six categories: Lung nodule or mass (1{,}096), Groundglass opacity (925), Atelectasis (412), High lung volume (301), Consolidation (284), and Pleural abnormality (118), as shown in Fig.~\ref{fig:cat_dist}. The disease distribution in our ReXGroundingCT subset is markedly imbalanced across our six categories. However the best performing model did not apply class-balancing or reweighting (e.g., oversampling, inverse-frequency sampling, focal/weighted losses), because the most prevalent class, \emph{lung nodule or mass}, is also the hardest to localize due to its small size, variable location, and sparse slice extent. This choice is supported by the disease-specific results in Table~\ref{tab:category_results} of the main paper, where XGLASS improves across all disease categories, and by the loss-function ablation in Sec.~\ref{sec:loss_ablation}, where reweighted objectives do not improve small-finding and overall retrieval performance.

\begin{figure*}[t]
    \centering
    \includegraphics[width=\linewidth]{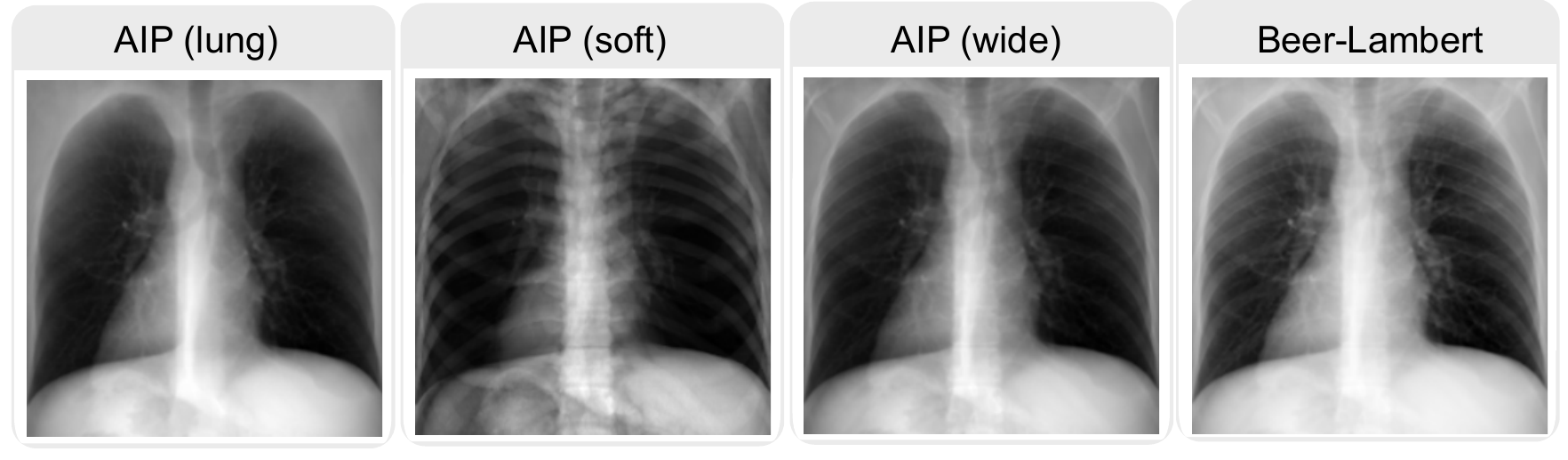}
     \caption{
    \textbf{CT volume projections in XGLASS.}
    Example coronal slice projections of the same chest CT under different projection methods and windowing settings.
    AIP (lung) emphasizes parenchymal findings and saturates mediastinal structures, AIP (soft) suppresses low-attenuation lung parenchyma, so bony structures dominate the appearance, AIP (wide) exposes a broader range of intensities, and the Beer–Lambert projection yields a more radiograph-like appearance with stronger contrast in high-attenuation structures.
    }
    \label{fig:ct_projection_gallery}
\end{figure*}

\subsection{CT Volume Projections}
For XGLASS, each 3D chest CT volume is converted into a 2D coronal intensity projection that lives in the same domain as our disease-conditioned importance maps, as explained in Section~3.2. Formally, we project along the anterior–posterior axis and aggregate attenuation values along each ray using an operator $A$ (Beer–Lambert or average-intensity), yielding a coronal image $P_A(u,v)$ for each volume. In the main experiments we focus on two projection families: (i) a physically motivated Beer-Lambert projection that mimics x-ray transmission, and (ii) average-intensity projections (AIP) combined with standard CT windowings. For AIP we consider three clinical window settings: a lung window, a soft-tissue window, and a wide window spanning a broad Hounsfield range. As discussed in Section 3.2 and ablated in Section 4.4.2, Beer-Lambert provides slightly better global ranking (AUC, Spearman), whereas AIP with a wide window offers the best early retrieval utility (Hit@K). Figure~\ref{fig:ct_projection_gallery} illustrates how a single CT scan appears under these different projection choices, highlighting how windowing and projection type emphasize complementary structures (e.g., parenchyma in the lung window vs.\ mediastinum and pleural surfaces under broader windows).

\subsection{Gaze-Prior Blending Strength}
\begin{figure*}[t]
    \centering
    \includegraphics[width=\linewidth]{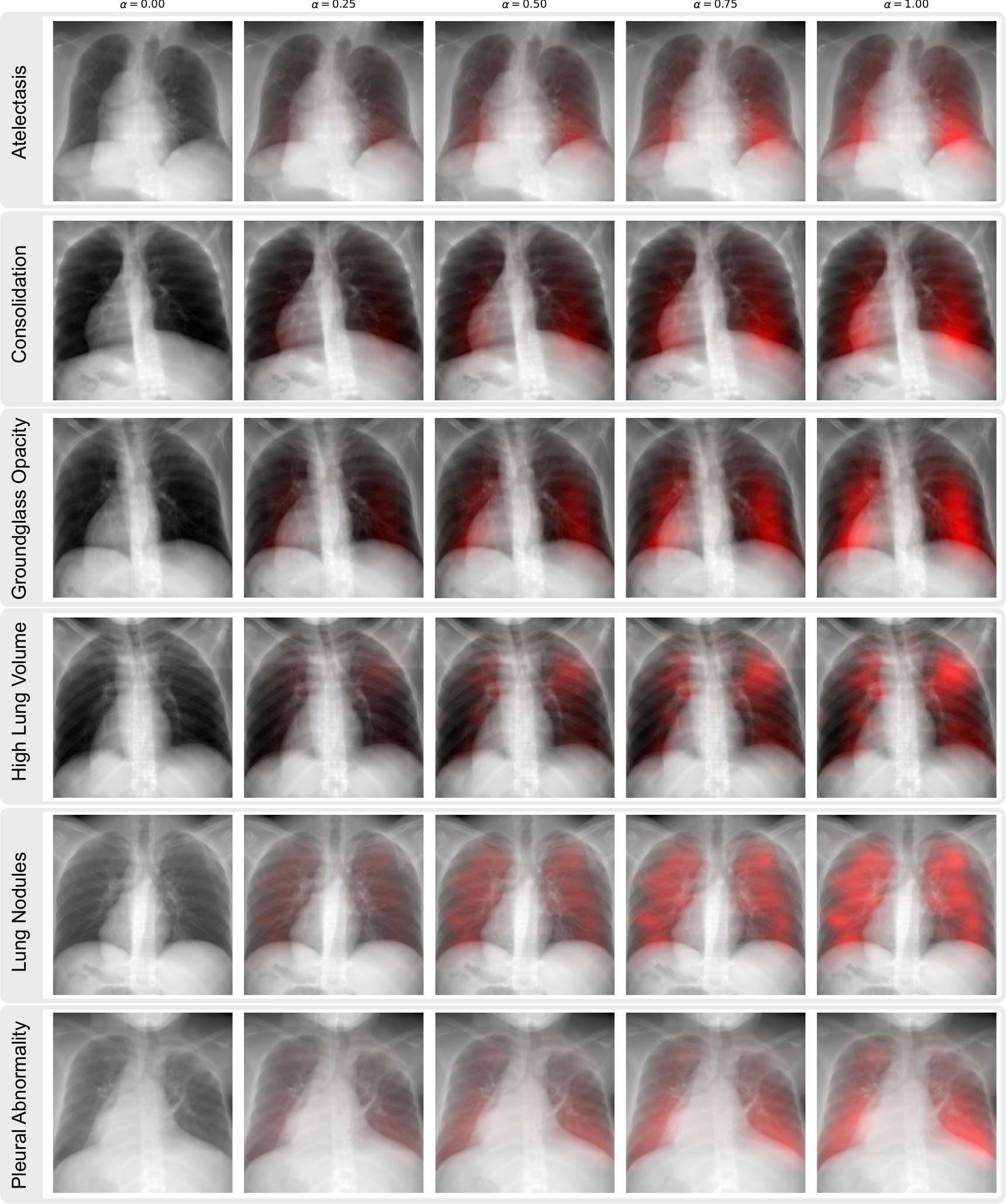}
    \caption{
   \textbf{ Effect of the gaze–heatmap blending strength $\alpha$} on the AIP-wide CT projections.
    Rows correspond to the six disease categories used in XGLASS (one representative CT per category), and columns sweep the blending coefficient from left to right: $\alpha \in \{0.0, 0.25, 0.5, 0.75, 1.0\}$.
    The leftmost column ($\alpha = 0$) shows the raw CT projection without gaze information, while increasing $\alpha$ progressively injects the REFLACX-derived gaze importance map into the red channel within the chest bounding box.
    }
    \label{fig:alpha_blend_panel}
\end{figure*}

In XGLASS, the 2D CT projection and the disease-specific REFLACX gaze importance map are combined by injecting the normalized gaze heatmap $H_d(u,v)$ into the red channel inside the chest bounding box using a blending coefficient $\alpha \in [0,1]$. Here, $\alpha=0$ corresponds to the CT-only input, whereas larger values increase the contribution of the disease-conditioned gaze prior in the fused representation. Figure~\ref{fig:alpha_blend_panel} illustrates the qualitative effect of varying $\alpha$ for representative CT studies. The quantitative test-set ablation of $\alpha$ is reported in Sec.~\ref{sec:gaze_fusion_ablation} of the main paper. The final value $\alpha{=}0.75$ was selected on the validation split, as described in Sec.~\ref{sec:supp_validation_selection}.

\section{Lung Nodule Baseline Comparison}
\label{sec:nodule_baselines}

\begin{table*}[t!]
\centering
\caption{\textbf{Lung-nodule-specific baseline comparison.}
Results on all 112 lung nodule/mass test cases, the most spatially sparse and variable finding category. XGLASS achieves the strongest Hit@K and AUC compared to the other 2D and 3D baselines, indicating more reliable early retrieval despite the difficulty of this disease.}
\label{tab:nodule_baselines}
\small
\begin{tabular*}{\textwidth}{@{\extracolsep{\fill}}lcccccccc}
\toprule
\textbf{Method} & \textbf{Hit@1} & \textbf{Hit@3} & \textbf{Hit@5} &
\textbf{Hit@10} & \textbf{Hit@20} & \textbf{AR@10} &
\textbf{AUC} & \textbf{Spearman ($\rho$)} \\
\midrule
Random
& 0.080 & 0.125 & 0.179 & 0.286 & 0.446 & 0.038 & 0.491 & -0.007 \\
Vote-MI~\cite{wang2024enhancing}
& 0.018 & 0.107 & 0.179 & 0.313 & 0.545 & 0.035 & 0.478 & -0.017 \\
BioMedCLIP~\cite{zhang2025multimodal}
& 0.080 & 0.143 & 0.214 & 0.286 & 0.455 & 0.062 & 0.585 & 0.093 \\
3D Merlin + MLP~\cite{blankemeier2024merlin}
& 0.080 & 0.107 & 0.143 & 0.259 & 0.420 & 0.057 & 0.688 & 0.125 \\
3D U-Net~\cite{cciccek20163d}
& 0.080 & 0.143 & 0.161 & 0.232 & 0.384 & \textbf{0.079} & 0.698 & \textbf{0.132} \\
XGLASS
& \textbf{0.107} & \textbf{0.205} & \textbf{0.286} &
\textbf{0.446} & \textbf{0.607} & 0.074 &
\textbf{0.722} & 0.129 \\
\bottomrule
\end{tabular*}
\end{table*}

Lung nodules/masses are the hardest evaluated category because they are small, sparse, and spatially variable. However, this relative weakness does not indicate near-random performance. Across all 112 nodule cases, XGLASS achieves the best Hit@K and AUC among all 2D and 3D baselines, showing that it more reliably retrieves at least one relevant slice early. The 3D U-Net obtains slightly higher AR@10 and Spearman's $\rho$, but performs substantially worse on Hit@5, Hit@10, and Hit@20 (Table~\ref{tab:nodule_baselines}).

\section{Hyperparameter Selection}
\label{sec:supp_validation_selection}

\begin{table}[h!]
\centering
\caption{\textbf{Validation-based hyperparameter selection.}
All values are computed on the validation split.
$\overline{\mathrm{Hit}}$ denotes the mean of Hit@5, Hit@10, and Hit@20.
Gray rows indicate the settings selected for the final model; bold values
indicate the best performance within each ablation.}
\label{tab:supp_val_selection}

\footnotesize
\setlength{\tabcolsep}{3.2pt}
\renewcommand{\arraystretch}{1.05}

\begin{tabular}{lcccccc}
\toprule
\textbf{Setting}
& \textbf{Hit@5}
& \textbf{Hit@10}
& \textbf{Hit@20}
& $\boldsymbol{\overline{\mathrm{Hit}}}$
& \textbf{AUC}
& $\boldsymbol{\rho}$ \\
\midrule
\multicolumn{7}{l}{\textit{Backbone}} \\
\rowcolor{gray!12}
\textbf{ViT-B/16}
    & \textbf{0.441} & \textbf{0.559} & \textbf{0.709}
    & \textbf{0.570} & 0.796 & 0.250 \\
RadioDino-s8
    & 0.435 & 0.539 & 0.644
    & 0.539 & \textbf{0.804} & \textbf{0.256} \\
Swin-Tiny
    & 0.405 & 0.464 & 0.578
    & 0.482 & 0.798 & 0.244 \\

\midrule
\multicolumn{7}{l}{\textit{CT projection}} \\
\rowcolor{gray!12}
\textbf{AIP-wide}
    & \textbf{0.441} & 0.559 & \textbf{0.709}
    & \textbf{0.570} & 0.796 & \textbf{0.250} \\
AIP-soft
    & 0.395 & \textbf{0.572} & 0.654
    & 0.540 & \textbf{0.798} & 0.245 \\
AIP-lung
    & 0.431 & 0.510 & 0.618
    & 0.520 & 0.794 & 0.242 \\
Beer-Lambert
    & 0.428 & 0.549 & 0.647
    & 0.541 & 0.796 & 0.246 \\

\midrule
\multicolumn{7}{l}{\textit{Gaze blending coefficient $\alpha$}} \\
0.00
    & 0.395 & 0.500 & 0.578 & 0.491 & 0.771 & 0.236 \\
0.25
    & 0.422 & 0.549 & 0.641 & 0.537 & 0.802 & 0.246 \\
0.50
    & 0.412 & 0.539 & 0.693 & 0.548 & 0.797 & 0.248 \\
\rowcolor{gray!12}
\textbf{0.75}
    & \textbf{0.441} & \textbf{0.559} & \textbf{0.709}
    & \textbf{0.570} & 0.796 & \textbf{0.250} \\
1.00
    & 0.428 & 0.513 & 0.657
    & 0.533 & \textbf{0.803} & 0.248 \\

\midrule
\multicolumn{7}{l}{\textit{Gaze-injection channel}} \\
\rowcolor{gray!12}
\textbf{R}
    & 0.441 & 0.559 & \textbf{0.709}
    & \textbf{0.570} & 0.796 & \textbf{0.250} \\
B
    & \textbf{0.451} & \textbf{0.575} & 0.683
    & \textbf{0.570} & 0.801 & 0.243 \\
R+B
    & 0.412 & 0.539 & 0.683
    & 0.545 & 0.804 & 0.248 \\
G
    & 0.415 & 0.536 & 0.650
    & 0.534 & \textbf{0.806} & 0.243 \\
G+B
    & 0.422 & 0.510 & 0.650
    & 0.527 & 0.798 & 0.241 \\
R+G
    & 0.435 & 0.520 & 0.624
    & 0.526 & 0.800 & 0.248 \\

\midrule
\multicolumn{7}{l}{\textit{Loss function}} \\
\rowcolor{gray!12}
\textbf{Area MSE}
    & \textbf{0.441} & 0.559 & \textbf{0.709}
    & \textbf{0.570} & 0.796 & 0.249 \\
Binary focal
    & 0.418 & \textbf{0.582} & 0.683
    & 0.561 & \textbf{0.804} & \textbf{0.250} \\
Slice-bal. MSE
    & 0.389 & 0.536 & 0.690
    & 0.538 & 0.726 & 0.229 \\
Size-wt. MSE
    & 0.405 & 0.520 & 0.670
    & 0.532 & 0.790 & 0.235 \\

\bottomrule
\end{tabular}
\end{table}

We selected the model's hyperparameters such as the visual backbone, CT projection, gaze-blending coefficient $\alpha$, loss function, and gaze-injection channel using the validation split. All optimization hyperparameters were kept fixed across configurations: AdamW with learning rates of $1\times10^{-4}$ for the regression head and $1\times10^{-5}$ for the ViT backbone, and a batch size of 16. Since XGLASS targets slice retrieval across different inspection budgets, selection was based on the mean of Hit@5, Hit@10, and Hit@20, denoted as $\overline{\mathrm{Hit}}$ in Table~\ref{tab:supp_val_selection}. We exclude Hit@1 and Hit@3 from the selection criterion because these very small retrieval budgets are more sensitive to single-slice rank changes. When $\overline{\mathrm{Hit}}$ was tied, AUC and Spearman's $\rho$ were considered as secondary ranking criteria. ViT-B/16, AIP-wide, and $\alpha{=}0.75$ achieved the highest $\overline{\mathrm{Hit}}$ within their respective ablations. For the gaze-injection channel, red and blue achieve the same $\overline{\mathrm{Hit}}$. Red yields a slightly higher Spearman correlation ($0.250$ vs.\ $0.243$) while differing only marginally in AUC ($0.796$ vs.\ $0.801$). Given the similar slice retrieval performance and the ranking objective of XGLASS, we therefore selected red-channel injection for the final model. In the loss-function ablation, unweighted MSE achieves a higher $\overline{\mathrm{Hit}}$ than binary focal loss and lesion-balanced MSE, while preserving competitive AUC and Spearman correlation; we therefore selected the unweighted MSE as the final training objective. The resulting configuration, ViT-B/16, AIP-wide, $\alpha{=}0.75$, red-channel gaze injection, and unweighted MSE, was fixed and used for the test-set ablations in Sec.~4.4 and Sec.~\ref{sec:extendedab}.

\section{Extended Ablations}
\label{sec:extendedab}

\subsection{Gaze-Injection Channel Combinations}
\label{sec:ablation_channel_injection}

We additionally evaluate every RGB-channel combination for injecting the gaze
prior at the selected blending strength $\alpha{=}0.75$ on the held-out test
set. Table~\ref{tab:ablation_channel_injection} shows that the two strongest
configurations both include the red channel: magenta ($R{+}B$) performs best
for AUC, Spearman correlation, Hit@1, and Hit@5, while red-only achieves
similar overall performance and the highest Hit@10 and Hit@20. Red-only was
selected on the validation split, where it tied with blue for
$\overline{\mathrm{Hit}}$ and achieved the higher Spearman correlation.
The test-set results further show that red-only remains particularly strong at
the larger retrieval budgets.

\begin{table}[h]
  \centering
  \caption{\textbf{Channel Injection Strategy Ablation.}
  Gaze-prior injection into individual RGB channels and their combinations.}
  \label{tab:ablation_channel_injection}
  \setlength{\tabcolsep}{3pt}
  \renewcommand{\arraystretch}{1.05}
  \small
  \resizebox{\columnwidth}{!}{%
  \begin{tabular}{lcccccc}
    \toprule
    \textbf{Channels} & \textbf{AUC} & \textbf{Spearman} &
    \textbf{Hit@1} & \textbf{Hit@5} & \textbf{Hit@10} & \textbf{Hit@20} \\
    \midrule
    R   & \underline{0.781} & \underline{0.224} & \underline{0.228} & \underline{0.421} & \textbf{0.563} & \textbf{0.693} \\
    G   & 0.754 & 0.212 & 0.193 & 0.345 & 0.459 & 0.582 \\
    B   & 0.761 & 0.209 & 0.190 & 0.418 & 0.519 & 0.655 \\
    R+G & 0.774 & 0.220 & 0.184 & 0.383 & 0.503 & 0.639 \\
    R+B & \textbf{0.784} & \textbf{0.226} & \textbf{0.241} & \textbf{0.443} & \underline{0.547} & \underline{0.661} \\
    G+B & 0.749 & 0.212 & 0.222 & 0.367 & 0.468 & 0.617 \\
    RGB & 0.761 & 0.196 & 0.192 & 0.364 & 0.465 & 0.594 \\
    \bottomrule
  \end{tabular}
  }
\end{table}

\subsection{XGLASS Backbone}
\label{sec:ablation_backbone_text}
Different backbones encode different inductive biases (e.g., local vs.\ global receptive fields, domain pretraining). We test a ViT-B/16 backbone (patch size $16$, input $224{\times}224$) with AugReg ImageNet-21k pretraining~\cite{dosovitskiy2020vit}, a Swin transformer with shifted windows (patch size $4$, window $7$) at $224{\times}224$, pretrained on ImageNet-22k~\cite{liu2021Swin}, and RadioDino-s8~\cite{ZEDDA2025110583}, a DINO-style self-supervised ViT pretrained on RadImageNet~\cite{mei2022radimagenet} as a domain-adapted medical backbone. Holding inputs fixed, this ablation isolates how architectural choices and pretraining affect the model’s ability to learn disease-aware slice-importance scores from our importance map enhanced CT intensity projections. The results in Table~\ref{tab:ablation_model} show that while Swin-Tiny attains the highest AUC and RadioDino-s8 slightly leads on Spearman's $\rho$, the validation-selected \textbf{ViT-B/16} backbone delivers the best diagnostically relevant slice retrieval performance, with higher Hit@K across $K{=}1,5,10$.

\begin{table}[h]
  \centering
  \caption{\textbf{Backbone Ablation.} Under AIP with wide window projection and $\alpha{=}0.75$.}
  \label{tab:ablation_model}
  \setlength{\tabcolsep}{2pt}
  \renewcommand{\arraystretch}{1.05}
  \small
  \resizebox{\columnwidth}{!}{%
  \begin{tabular}{lcccccc}
    \toprule
    \textbf{Backbone} & \textbf{AUC} & \textbf{Spear.} & \textbf{Hit@1} & \textbf{Hit@5} & \textbf{Hit@10}  \\
    \midrule
    RadioDino-s8  & 0.777 & \textbf{0.227} & 0.180 & \underline{0.361} & \underline{0.487} \\
    Swin-Tiny     & \textbf{0.784} & 0.222 & \underline{0.203} & 0.358 & 0.472  \\
    \textbf{ViT-B/16}      & \underline{0.781} & \underline{0.224} & \textbf{0.228} & \textbf{0.421} & \textbf{0.563}\\
    
    \bottomrule
  \end{tabular}
  }
\end{table}

\subsection{Gaze-Only and CT-Derived Prior Ablations}
\label{sec:gaze}

We further analyze the role of disease-conditioned priors in XGLASS by comparing transferred CXR gaze priors with empirical CT abnormality-location priors. First, we evaluate the CXR gaze prior alone by averaging each disease-specific gaze map along the left--right axis, interpolating the resulting superior--inferior profile to 240 axial positions, and ranking slices by profile intensity. Second, we construct an empirical CT location prior using only train+val labels by averaging slice-presence vectors for each disease category and applying the resulting fixed category-specific prior to all corresponding test cases. Third, we replace the gaze map in XGLASS with this CT-derived prior converted into a synthetic heatmap and fuse it with the CT projection in the same way as the gaze heatmap. 
Since this CT prior is computed from train+val ground-truth slice-presence targets from the same CT dataset distribution, it represents an empirical average lesion-location prior for each abnormality. In contrast, the gaze prior is transferred from CXR eye-tracking and does not use CT lesion-location statistics; instead, it captures radiologist visual search patterns that include both observed abnormality regions and plausible candidate regions inspected for each finding. In contrast, the CT-derived prior is less reliable when abnormalities occur in rare locations, whereas the broader radiologist search patterns encoded by the gaze prior can still overlap these regions, supporting the use of gaze over CT derived location priors. Table~\ref{tab:gaze_prior_only} shows that CT location statistics are informative: the CT location prior alone achieves the best Hit@1 and Spearman correlation. However, the full gaze-guided XGLASS model achieves the best AUC and performs substantially better for utility-centric top-$K$ retrieval metrics, improving Hit@10 from $0.453$ to $0.563$ and Hit@20 from $0.522$ to $0.693$ compared with the CT location prior alone. Injecting the CT-derived prior into XGLASS also improves over CT-only input at Hit@10 ($0.535$ vs. $0.468$). Nevertheless, gaze-guided XGLASS remains stronger at Hit@10 and Hit@20, indicating that the transferred radiologist gaze prior provides complementary information beyond average CT lesion-frequency statistics, which improves the slice selection performance especially for higher $K$. We additionally stratify test samples by their typicality under the train+val CT location prior, defined as the average CT-prior score over the ground-truth positive slices of each test sample. In low-typicality cases, where the empirical CT prior assigns low probability to the true lesion region, the CT location prior drops sharply to Hit@10 of $0.151$. In contrast, the CXR gaze prior alone remains stronger at Hit@10 of $0.330$, while full XGLASS reaches Hit@10 of $0.358$ and Hit@20 of $0.528$. This suggests that empirical CT priors primarily capture common dataset-specific lesion locations, whereas radiologist gaze priors cover broader disease-specific search regions. Combined with patient-specific CT projection evidence, this leads to more robust disease-conditioned slice prioritization than static location priors alone.

\begin{table}[h]
\centering
\caption{\textbf{Prior-Only and Prior-Injection Ablations.}
Comparison of static CXR gaze priors, empirical CT location priors, CT-only XGLASS, CT-prior injection, and full gaze-guided XGLASS.}
\label{tab:gaze_prior_only}
\setlength{\tabcolsep}{2.5pt}
\renewcommand{\arraystretch}{1.05}
\small
\resizebox{\columnwidth}{!}{%
\begin{tabular}{lcccccc}
\toprule
\textbf{Method} & \textbf{AUC} & \textbf{Spear.} &
\textbf{Hit@1} & \textbf{Hit@5} & \textbf{Hit@10} & \textbf{Hit@20} \\
\midrule
Gaze prior only & 0.638 & 0.163 & 0.158 & 0.209 & 0.263 & 0.358 \\
CT location prior only & 0.734 & \textbf{0.229} & \textbf{0.250} & 0.329 & 0.453 & 0.522 \\
XGLASS w/o gaze cues & \underline{0.768} & 0.212 & 0.199 & 0.342 & 0.468 & 0.619 \\
XGLASS + CT loc. prior & 0.722 & 0.199 & 0.199 & \textbf{0.424} & \underline{0.535} & \underline{0.658} \\
XGLASS w/ gaze cues & \textbf{0.781} & \underline{0.224} &
\underline{0.228} & \underline{0.421} &
\textbf{0.563} & \textbf{0.693} \\
\bottomrule
\end{tabular}%
}
\end{table}

\subsection{Robustness to Gaze-Prior Perturbations}
\label{sec:gaze_robustness}

To test whether XGLASS depends on exact pixel-level alignment of the transferred gaze prior, we first measured coarse anatomical alignment between CXR-derived priors and CT projections using lung and heart layouts, obtaining a lungs+heart centroid error of 16.2 px and an average symmetric surface distance of 11.6 px; for lungs alone, the corresponding errors were 8.8 px and 8.0 px. We then perturb the disease-conditioned heatmap at test time while keeping the CT projection unchanged. Table~\ref{tab:gaze_robustness} shows that random spatial shifts, additive noise, and left--right heatmap flips only mildly affect performance, supporting that XGLASS uses the gaze map as a broad anatomical search prior, that is robust to minor perturbations, rather than an exact lesion-location mask.

\begin{table}[h]
\centering
\caption{\textbf{Robustness to Gaze-Prior Perturbations.}
Test-time perturbations of the disease-conditioned gaze map.}
\label{tab:gaze_robustness}
\setlength{\tabcolsep}{3pt}
\renewcommand{\arraystretch}{1.05}
\small
\resizebox{\columnwidth}{!}{%
\begin{tabular}{lccccc}
\toprule
\textbf{Condition} & \textbf{AUC} & \textbf{Spear.} & \textbf{Hit@1} & \textbf{Hit@5} & \textbf{Hit@10} \\
\midrule
Baseline & \underline{0.781} & \textbf{0.224} & \textbf{0.228} & \underline{0.421} & \textbf{0.563} \\
Random shift $\pm16$ px & 0.778 & \underline{0.221} & 0.206 & 0.405 & 0.528 \\
Noise $\sigma{=}0.1$ & 0.775 & 0.220 & \underline{0.215} & \textbf{0.427} & \underline{0.557} \\
Left--right flip & \textbf{0.786} & \textbf{0.224} & 0.203 & 0.396 & \textbf{0.563} \\
\bottomrule
\end{tabular}}
\end{table}

\subsection{Loss Function Ablation}
\label{sec:loss_ablation}
Abnormalities vary substantially in spatial extent, e.g., consolidations often cover larger regions than lung nodules. We therefore compare four training objectives while keeping all other settings fixed: MSE regresses the normalized slice-area target, Focal treats each slice as positive or negative and emphasizes difficult cases, Slice-bal. MSE gives equal aggregate weight to positive and negative slices within each CT, preventing scans with many positive slices from dominating the loss, and Size-wt. MSE inversely weights positive-slice errors by the total lesion size in each CT, reducing the contribution of spatially extensive lesions. Table~\ref{tab:ablation_loss_function} shows that MSE gives the strongest overall retrieval-ranking trade-off, with the best Hit@1 and Hit@10 and tied-best Hit@5. We further stratify the same ablation by finding extent. Test cases are split into small, medium, and large tertiles by total ground-truth slice-area mass, corresponding to average extents of 5.6, 14.7, and 52.4 positive slices, respectively.

\begin{table}[h]
  \centering
  \caption{\textbf{Loss-function ablation.} We compare Area MSE with binary and lesion-size-aware alternatives.}
  \label{tab:ablation_loss_function}
  \setlength{\tabcolsep}{2pt}
  \renewcommand{\arraystretch}{1.05}
  \small
  \resizebox{\columnwidth}{!}{%
  \begin{tabular}{lccccc}
    \toprule
    \textbf{Loss} & \textbf{AUC} & \textbf{Spear.} & \textbf{Hit@1} & \textbf{Hit@5} & \textbf{Hit@10} \\
    \midrule
    \textbf{MSE}
      & 0.781 & \textbf{0.224} & \textbf{0.228} & \textbf{0.421} & \textbf{0.563} \\
    Focal
      & \textbf{0.790} & 0.218 & 0.180 & 0.392 & 0.516 \\
    Slice-bal. MSE
      & 0.709 & 0.202 & 0.206 & \textbf{0.421} & \underline{0.519} \\
    Size-wt. MSE
      & \underline{0.785} & \underline{0.221} & \underline{0.215} & \underline{0.411} & \underline{0.519} \\
    \bottomrule
  \end{tabular}
  }
\end{table}

\noindent Table~\ref{tab:ablation_loss_by_size} shows that MSE is strongest for small sparse findings across retrieval metrics, whereas Focal and Slice-bal. MSE mainly improve medium or large findings. Although the reweighted objectives were intended to improve small-lesion retrieval, they appear to amplify the instability of very sparse positive targets: for small findings, only a few slices are labeled relevant and the corresponding projection evidence is often weak, so upweighting these slices can emphasize case-specific noise. In contrast, MSE preserves the continuous slice-area target and provides a smoother ranking signal, making it the most stable default objective.

\begin{table}[h]
\centering
\caption{\textbf{Loss retrieval metrics by finding size.}
Test cases are grouped into tertiles by total ground-truth slice-area mass. We report retrieval metrics within each size group.}
\label{tab:ablation_loss_by_size}
\small
\setlength{\tabcolsep}{3pt}
\renewcommand{\arraystretch}{1.05}
\resizebox{\columnwidth}{!}{%
\begin{tabular}{llccccc}
\toprule
\textbf{Loss} & \textbf{Size} & \textbf{Hit@1} & \textbf{Hit@3} & \textbf{Hit@5} & \textbf{Hit@10} & \textbf{AR@10} \\
\midrule
\multirow{3}{*}{MSE}
& Small  & \textbf{0.057} & \textbf{0.123} & \textbf{0.189} & \textbf{0.340} & \textbf{0.103} \\
& Medium & \underline{0.114} & 0.276 & 0.352 & \underline{0.543} & 0.079 \\
& Large  & \textbf{0.514} & 0.619 & \underline{0.724} & \underline{0.810} & \underline{0.090} \\
\midrule
\multirow{3}{*}{Focal}
& Small  & 0.019 & 0.075 & 0.104 & 0.226 & 0.065 \\
& Medium & \textbf{0.162} & \textbf{0.352} & \textbf{0.476} & \textbf{0.581} & \textbf{0.108} \\
& Large  & 0.438 & 0.600 & 0.667 & 0.752 & \textbf{0.090} \\
\midrule
\multirow{3}{*}{Slice-bal. MSE}
& Small  & \underline{0.047} & \underline{0.113} & 0.142 & 0.208 & 0.073 \\
& Medium & 0.105 & 0.257 & 0.352 & 0.486 & 0.077 \\
& Large  & 0.467 & \textbf{0.657} & \textbf{0.771} & \textbf{0.867} & 0.087 \\
\midrule
\multirow{3}{*}{Size-wt. MSE}
& Small  & 0.028 & 0.104 & \underline{0.151} & \underline{0.264} & \underline{0.085} \\
& Medium & \underline{0.114} & \underline{0.257} & \underline{0.362} & 0.505 & \underline{0.087} \\
& Large  & \underline{0.505} & \underline{0.648} & \underline{0.724} & 0.790 & 0.089 \\
\bottomrule
\end{tabular}
}
\end{table}

\section{Additional Implementation Details}
\label{sec:baseline_downstream_details}

\subsection{BioMedCLIP Slice Retrieval}
\label{sec:biomedclip_details}

We use BioMedCLIP~\cite{zhang2025multimodal} as a zero-shot slice-wise image--text retrieval baseline. We follow its official evaluation pipeline by rendering each axial CT slice with a wide window (center $0$ HU, width $2000$ HU), converted to RGB, and encoded independently. Finding names are normalized where appropriate (e.g., ``lung nodule or mass'' to ``pulmonary nodule'' and ``groundglass opacity'' to ``ground-glass opacity''). To reduce sensitivity to a single prompt formulation, we average the cosine similarities obtained from the following five templates, where \texttt{\{finding\}} denotes the normalized finding name:

\begin{enumerate}
    \item \texttt{The radiological finding is \{finding\}.}
    \item \texttt{This axial CT image shows \{finding\}.}
    \item \texttt{The CT slice is positive for \{finding\}.}
    \item \texttt{There is \{finding\} in this CT slice.}
    \item \texttt{A finding of \{finding\} is present on this CT image.}
\end{enumerate}

\noindent Slices are ranked in descending order of the ensemble-averaged similarity. For the downstream binary-screening analysis, one global similarity threshold ($0.3751$) is selected on the all-slice BioMedCLIP baseline by maximizing F1 and then applied unchanged to every XGLASS Top-$K$ subset.

\begin{figure*}[t]
    \centering
    \includegraphics[width=\linewidth]{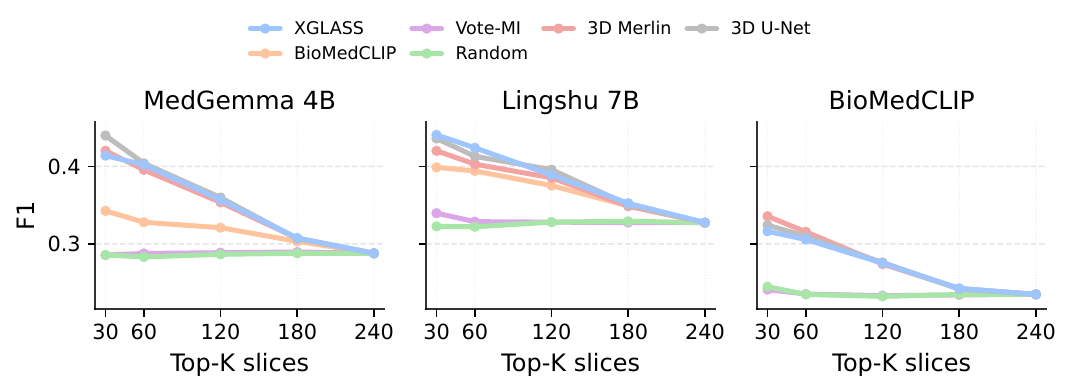}
    \caption{
\textbf{Matched-budget downstream screening across slice selectors.}
    For each CT--finding pair, different selectors choose the Top-$K$ inspected axial slices under the same slice budget.
    XGLASS is compared with BioMedCLIP retrieval, Vote-MI, 3D u-Net, 3D Merlin and Random selection across MedGemma 4B, Lingshu 7B, and BioMedCLIP downstream screening.
    }
    \label{fig:matched_budget_downstream}
\end{figure*}

\subsection{3D Baseline Implementation Details}
\label{sec:3d_baseline_details}

For the 3D Merlin + MLP baseline, we follow the CT preparation used by Merlin~\cite{blankemeier2024merlin}: full CT volumes are resampled to $1.5{\times}1.5{\times}3.0$\,mm spacing and resized to $160{\times}224{\times}224$ before encoding. The Merlin encoder is kept frozen and used to precompute one 2048-dimensional volume embedding per CT. We then train a disease-conditioned MLP on ReXGroundingCT to regress the same 240-dimensional slice-relevance target as XGLASS. The MLP uses a 32-dimensional disease embedding concatenated with the Merlin feature, followed by two 512-unit hidden layers with ReLU activations and dropout $0.1$, and a final linear layer producing 240 slice scores. The MLP is trained with AdamW, learning rate $10^{-3}$, weight decay $10^{-4}$, batch size 64, and the best checkpoint is selected on the validation loss. Including embedding precomputation, the full Merlin baseline required approximately 6 hours on an A100 GPU. For the \textbf{3D U-Net} baseline, we train a compact 3D U-Net from scratch on ReXGroundingCT. Full CT volumes are resampled to $1.25$\,mm isotropic spacing and resized to $160{\times}224{\times}224$. The encoder uses 3D convolutional blocks with base width 16 and instance normalization. To obtain slice scores, the final feature maps are averaged over the in-plane dimensions, interpolated to 240 axial positions, concatenated with a 32-dimensional disease embedding, and passed through a slice-regression head. The model is trained with AdamW, learning rate $10^{-4}$, weight decay $10^{-4}$, batch size 16, dropout $0.1$, and the best checkpoint is selected on the validation loss and Training to the selected checkpoint required approximately 22 hours on an A100 GPU. Both 3D baselines are therefore substantially more compute-intensive to train than XGLASS, which trains in under 10 minutes on a single RTX 4080.

\subsection{Slice-wise Medical VLM Screening}
\label{sec:vlm_screening_details}
We evaluate \texttt{google/medgemma-4b-it}~\cite{sellergren2025medgemma} and \texttt{lingshu-medical-mllm/Lingshu-7B}~\cite{xu2025lingshu} using the same disease-conditioned binary screening task. Motivated by CT vision-language transfer work that maps complementary HU windows to RGB channels to expose different tissue contrasts~\cite{minoccheri2026representation}, the main-paper experiments use a pseudo-RGB representation with wide/lung, soft-tissue, and mediastinal/fluid windows assigned to separate channels. As an input-rendering sensitivity analysis, we additionally evaluate a standard wide-window representation ($-1024$ to $1024$ HU) and replicated across all RGB channels. This alternative setting shows the same trend: XGLASS Top-$K$ preselection improves F1 over all-slice screening, with the strongest gains at small budgets (e.g., Top-30 F1 of 0.414 for MedGemma and 0.441 for Lingshu). Thus, the downstream benefit of XGLASS is consistent across CT slice renderings. Both models receive the same system and user instructions:

\begin{quote}
\small
\textbf{System:} You are an expert thoracic radiologist. Evaluate only the queried finding on one axial chest CT slice.

\textbf{User:} Are there clear visual signs of \texttt{\{finding\}} in this axial CT slice? Ignore other abnormalities. If the queried finding is not visible in this slice, answer no. Return only one short line, with no explanation: \texttt{Answer: yes} or \texttt{Answer: no}.
\end{quote}

\noindent We use deterministic decoding and parse the response as yes, no, or unknown; no unknown responses occurred in the reported MedGemma and Lingshu evaluation. MedGemma is evaluated in bfloat16 without quantization, while Lingshu is loaded with 4-bit NF4 quantization and bfloat16 computation due to memory requirements. Predictions are generated independently for every slice and cached, allowing identical predictions to be evaluated on either all 240 slices or XGLASS-selected Top-$K$ subsets. Precision, recall, and F1 are computed only over the inspected slices.

\subsection{Downstream Slice Selection Comparison}
\label{sec:matched_budget_downstream}

We further analyze how slice selection affects downstream screening when the number of inspected slices is fixed. For each CT--finding pair, we vary only the ranking used to choose the Top-$K$ inspected slices, while keeping the downstream slice-wise predictions unchanged. This isolates the effect of slice selection from stochasticity or repeated downstream model calls. We compare XGLASS with Random selection, Vote-MI, BioMedCLIP retrieval, and the two 3D-native selectors, 3D Merlin + MLP and 3D U-Net. BioMedCLIP retrieval is evaluated as a selector for MedGemma and Lingshu, but omitted from the BioMedCLIP downstream panel to avoid selecting and classifying slices with the same image--text similarity model. Figure~\ref{fig:matched_budget_downstream} shows that selector choice strongly affects low-budget screening, especially at $K{=}30$ and $K{=}60$. XGLASS consistently outperforms the other 2D selectors under matched slice budgets, while the 3D selectors are competitive and sometimes achieve the highest downstream F1 for specific model--budget combinations. Notably, XGLASS remains close to these volumetric selectors despite using only a lightweight 2D projection, and outperforms them for Lingshu at $K{=}30$, $60$, and $180$. This suggests that the disease-conditioned XGLASS ranking transfers effectively to downstream screening, narrowing much of the gap to more expensive 3D-native selection.

\subsection{Downstream Computational Cost}
\label{sec:compute_details}
Applying the downstream models to all 240 slices of one CT requires approximately $1.40$\,s for BioMedCLIP, $154.91$\,s for MedGemma, and $267.83$\,s for Lingshu on a single NVIDIA RTX 4080. XGLASS requires only $0.025$\,s per finding ($0.150$\,s per CT) to rank the complete volume and allows the downstream model to inspect a smaller, disease-enriched subset. At Top-30, including XGLASS preselection, the corresponding per-CT runtimes decrease to approximately $0.20$\,s, $19.39$\,s, and $33.51$\,s, respectively. Thus, XGLASS reduces downstream runtime by $85.7$--$87.5\%$ while simultaneously improving F1 at the smaller slice budget, showing that preselection improves both computational efficiency and downstream screening performance.

\end{document}